\documentclass[final,5p,times,twocolumn]{elsarticle}

\usepackage{amsfonts}
\usepackage[latin1]{inputenc}

\usepackage{amssymb}
\usepackage{times}
\usepackage{graphicx} 
\usepackage{epstopdf,epsfig} 

\usepackage{multirow}
\usepackage{natbib}

\usepackage[ruled,vlined]{algorithm2e}

\usepackage{amsthm}
\usepackage{amsmath}

\journal{Neurocomputing}

\begin{document}
\def \xx {\mathbf{x}}
\def \ww {\mathbf{w}}
\def \X  {\mathcal{X}}
\def \Y  {\mathcal{Y}}
\def \R  {\mathcal{R}}
\def \U  {\mathcal{U}}
\def \F  {\mathcal{F}}
\def \ee {\mathbf{e}}
\def \alphaa  {\mbox{\boldmath $\alpha$}}
\def \muu  {\mbox{\boldmath $\mu$}}
\def \lambdaa  {\mbox{\boldmath $\lambda$}}
\def \o  {\mbox{\boldmath $1$}}

\newtheorem{thm}{Theorem}
\newtheorem{lem}[thm]{Lemma}
\newdefinition{rmk}{Remark}
\newdefinition{defn}{Definition}
\newproof{pf}{Proof}
\newproof{pot1}{Proof of Theorem \ref{thm1}}

\begin{frontmatter}



\title{Making Risk Minimization Tolerant to Label Noise}
\author[iisc]{Aritra Ghosh}
\ead{aritraghosh.iem@gmail.com}
\author[ge]{Naresh~Manwani\corref{cor1}}
\ead{nareshmanwani@gmail.com}
\author[iisc]{P.~S.~Sastry}
\ead{sastry@ee.iisc.ernet.in}

\cortext[cor1]{Corresponding author}
\address[iisc]{Department of Electrical Engineering, Indian Institute of Science, Bangalore - 560012, India}
\address[ge]{GE Global Research, John F. Welch Technology Centre, \# 122, EPIP Phase 2, Whitefield Road, Hoodi Village, Bangalore- 560066, India}

\begin{abstract}
In many applications, the training data, from which one needs to learn a classifier, is corrupted with label noise. Many standard algorithms such as SVM perform poorly in presence of label noise. In this paper we investigate the robustness of risk minimization to label noise. We prove a sufficient condition on a loss function for the risk minimization under that loss to be tolerant to uniform label noise. We show that the $0-1$ loss, sigmoid loss, ramp loss and probit loss satisfy this condition though none of the standard convex loss functions satisfy it. We also prove that, by choosing a sufficiently large value of a parameter in the loss function, the sigmoid loss, ramp loss and probit loss can be made tolerant to non-uniform label noise also if we can assume the classes to be separable under noise-free data distribution. Through extensive empirical studies, we show that risk minimization under the $0-1$ loss, the sigmoid loss and the ramp loss has much better robustness to label noise when compared to the SVM algorithm.

\end{abstract}


\end{frontmatter}

\section{Introduction}
\label{Section:Introduction}

In a classifier learning problem we are given training data and when the class labels in the training data may be incorrect 
(or noise-corrupted), we refer to it as label noise.  
Learning classifiers in the presence of label noise is a classical problem in machine learning~\citep{frenay2013classification}. 
This challenging problem has become more relevant in recent times due to the 
current applications of Machine Learning. 
In many of the web based applications, the labeled data is essentially obtained through user feedback or user labeling. This leads 
to data with label noise because of a lot of variability among different users while labeling and also due to the inevitable 
human errors. In traditional pattern recognition problems also, we need to tackle label noise. 
For example,  overlapping class-conditional densities give rise to 
training data with label noise. This is because we can always view data generated from such densities as data that is 
originally  classified according to, say, Bayes optimal classifier and then subjected to (non-uniform) label noise before 
being given to the learning algorithm. 
Feature measurement errors can also lead to label noise in the training data.
  
In this paper, we discuss methods for learning classifiers that are robust to label noise. Specifically we consider the 
risk minimization strategy which is a generic method for learning classifiers. We focus on the issue of making risk minimization 
robust to label noise.  

Risk minimization is one of the popular strategies for learning classifiers from training 
data \citep{Hau1992,DGL1996}.\footnote{Risk minimization strategy is briefly discussed in Section~\ref{sec:risk-minimization}.} 
Many of the standard approaches for learning classifiers (such as Bayes classifier, Neural Network or SVM based classifier etc.) 
can be viewed as (empirical) risk minimization under a suitable loss function. 
The Bayes classifier minimizes risk under the $0-1$ loss function. 
One would like to minimize risk under $0-1$ loss as it minimizes probability of mis-classification. 
However, in general, minimizing risk under $0-1$ loss  
 is computationally hard because it gives rise to a non-convex and non-smooth optimization problem. 
Hence many convex loss functions are proposed to make the risk minimization efficient.  
Square loss (used in feed-forward neural networks), 
 Hinge loss (used in SVM), log-loss (used in logistic regression) and exponential loss (used in boosting)  
are some common examples of such convex loss functions. 
Many such convex loss functions are shown to be  {\em classification calibrated}; that is,   
low risk under these losses implies low risk under $0-1$ loss \citep{bjm-ccrb-05}.
However, these results do not say anything about the robustness of such risk minimization algorithms to label noise. 
In this paper we present some interesting theoretical results on when risk minimization can be robust to label noise. 

A learning algorithm can be said to be robust to label noise if 
the classifier learnt using noisy data and noise free data, both have same
classification accuracy on  noise-free test data~\citep{DBLP:journals/tcyb/ManwaniS13}. 
In \citet{DBLP:journals/tcyb/ManwaniS13}, it is shown that risk minimization under $0-1$ loss is tolerant to 
uniform noise (with noise rate less than 50\%). It is also tolerant to non-uniform noise under some additional 
 conditions. 
It is also shown in~\citep{DBLP:journals/tcyb/ManwaniS13} through counter-examples that 
 risk minimization under many of the standard convex loss functions such as hinge loss, 
log loss or exponential loss, is 
not noise-tolerant even under uniform noise. 

In this paper, we extend the above theoretical analysis. We provide some sufficient conditions on a loss function so that
 risk minimization with that loss function becomes noise tolerant under uniform and non-uniform label noise. 
While $0-1$ loss satisfies these, none of the standard convex loss functions satisfy the conditions. 
We also show that some of the non-convex loss functions such as sigmoid loss, ramp loss and probit loss satisfy the 
sufficiency conditions. Our results show that risk minimization under these loss functions is tolerant to uniform noise 
and that it is also tolerant to non-uniform noise  if the Bayes risk (under noise-free data) is zero and if one parameter 
in the loss function is properly chosen.  Hence we propose that risk minimization using sigmoid or ramp loss 
(which can be viewed as continuous but non-convex approximations to $0-1$ loss) would result in  
 learning methods that are robust to label noise. Through extensive empirical studies, we show that such risk minimization 
has good robustness to label noise. 

The rest of the paper is organized as follows. In Section~\ref{sec:prior-work}, we provide a brief review 
of methods for tackling label noise and then summarize the contributions of this paper. 
In Section~\ref{sec:problem} we define the notion of noise tolerance of a learning algorithm and formally 
state our problem. In this section we also provide a brief overview of the general risk minimization strategy. 
Section~\ref{sec:theory} contains all our theoretical results. We present simulation results on both 
synthetically generated data as well as  on some benchmark data sets in Section~\ref{sec:experiments}. Some concluding remarks are 
presented in Section~\ref{sec:conclusions}.

\section{Prior Work}
\label{sec:prior-work}

Learning in presence of noise is a long standing problem in machine learning. 
It has been approached from many different directions. 
A detailed survey of these approaches is given in \citet{frenay2013classification}.

In a recent study, Nettleton et al. present an extensive empirical investigation of robustness of many standard classifier
learning methods to noise in training data~\citep{NOF2010}. They showed that the Naive Bayes classifier has the best
noise tolerance properties. We comment more about this after presenting our theoretical results.

In general, when there is label noise, there are two broad approaches to the 
problem of learning a classifier. 
In the first set of approaches, data is preprocessed to clean the noisy points and then a classifier is learnt using standard algorithms. 
In the second set of approaches, the learning algorithm itself is designed in such a way that 
the label noise does not affect the algorithm.  
We call these approaches {\em inherently noise tolerant}. We briefly discuss these two broad approaches below.

\subsection{Data Cleaning Based Approaches} 

These approaches rely on guessing points which are corrupted by label noise. 
Once these points are identified, they can be either filtered out or their labels suitably altered. 
Several heuristics have been used to guess such noisy points.

For example, it is reasonable to assume that the class label of a point which is situated deep inside the class region of a class should match with
the class labels of its nearest neighbors. Thus, mismatch of the class label of a point with most of its nearest neighbors
can be used as a heuristic to decide whether a point is noisy or not \citep{fine1999noise}. This method of guessing noisy points may
not work near the classification boundary. The performance of this heuristic also depends on the number of nearest neighbors used.

Another heuristic is that, in general, noisy points are tough to classify correctly. Thus, when we learn multiple classifiers
using the noisy data, many of the classifiers may disagree on the class label of the noisy points.
This heuristic has also been used to identify noisy points \citep{Angelova:2005:PTS:1068507.1068955,
Brodley99identifyingmislabeled,Zhu:2003}. Decision tree pruning \citep{John95robustdecision},
distance of a point to the centroid of its own class \citep{daza2007algorithm},
points achieving weights higher than a threshold in boosting algorithm \citep{Karmaker:2006},
margin of the learnt classifier \citep{Sariel2007}
are some other heuristics which have been used to identify the noisy examples.

As is easy to see, the performance of such heuristics depend on the nature of label noise. 
There is no single approach for identifying noisy points which can work for all problems.
While each of the above heuristics has certain advantages, none of them are universally applicable. 
A non-noisy points can be detected as noisy point and vice-versa under any of these heuristics. 
This could eventually increase the overall noise level
in the training data. Moreover, removal of the noisy points from the
training data may lead to loosing important information about the classification boundary \citep{journals/pr/BouveyronG09}.

\subsection{Inherently Noise Tolerant Approaches}

These approaches do not do any preprocessing of the data; but the algorithm is designed in such a 
way that its output is not affected much by the label noise in the training data. 

Perceptron algorithm, which is the simplest algorithm for learning linear classifiers, is modified in several ways to make it robust to
the label noise \citep{Khardon:2007:NTV:1248659.1248667}.
Noisy points can frequently participate in updating the hyperplane parameters in the Perceptron algorithm, as noisy points are
tough to be correctly classified.
Thus, allowing a negative margin around the classification boundary can avoid frequent hyperplane updates caused due
to the misclassifications with small margin. Putting an upper bound on the number of mistakes allowed for any example
also controls the effect of label noise \citep{Khardon:2007:NTV:1248659.1248667}.
Similar techniques have been employed to improve Adaboost algorithm against noisy points.
Overfitting problem in Adaboost, caused due to the label noise, can be controlled by introducing a prior on 
weights which can punish large weights \citep{ROM1999}.
In boosting algorithms, making the coefficients of each of the base classifiers input-dependent, 
also controls the exponential growth of weights due to noise \citep{jin2003new}.
SVM can  be made robust to label noise by modifying the kernel matrix \citep{DBLP:journals/jmlr/BiggioNL11}. 
All these approaches are based on heuristics and work well in some  cases. 
However, for most of these approaches, there are no provable guarantees of noise tolerance. 

Noise tolerant learning has also been approached from the point of view of efficient probably approximately correct (PAC) learnability.
 By efficiency, we mean
polynomial time learnability. \citet{Kearns:98} proposed a PAC learning algorithm for learning 
 under label noise using statistical queries. However, the specific 
statistics that are calculated from the training data are problem-specific. 
PAC learning of the linear threshold functions is, in general,  NP-hard \citep{Hoffgen:1992}.
However,
linear threshold functions are efficiently PAC learnable under uniform noise if the noise-free data is 
linearly separable with appropriate large margin \citep{Bylander94}. For the same problem, 
\citet{Blum:1996} present a method to PAC-learn in presence of uniform
label noise without requiring the large margin condition. But the final
classifier is a decision list of linear threshold functions.
\citet{cohen1997learning} proposed an ellipsoid algorithm which efficiently PAC learns linear classifiers
under uniform label noise. This result is generalized further for class conditional label noise \citep{stempfel2007learning}.
(Under class conditional noise model, the probability of a label being corrupted is same for all examples of one class 
though different classes can have different noise rates). 
All these results are given for linear classifiers and for uniform label noise. 
There are no efficient PAC learnability results under non-uniform label noise. 

Recently \citet{conf/colt/ScottBH13} proposed a method of estimating Type~{\rm 1} and Type~{\rm 2} 
error rates of any specific classifier under the noise-free distribution given only the noisy  training data. 
This is for the case of a 2-class problem where the training data is corrupted with class conditional label noise.  
They used the concept of mutually irreducible distributions and  showed that 
such an estimation is possible if the noise-free class conditional distributions are mutually irreducible. 
This estimation strategy can be used to get a robust method of learning classifiers under class-conditional noise. 
In another recent method, \citet{NIPS2013_5073} propose risk minimization under a specially constructed surrogate 
loss function as a method of learning classifiers that is robust to class conditional label noise. 
Given any loss function, they propose a method to construct a new loss function. 
They show that the risk under this new loss for noisy data is same
as the risk under the original loss for noise free data. 
The construction of the new loss function needs information of noise rates which is to be estimated from data. 
Similar results are also presented in~\citep{Stempfel:2009}. 

\citet{DBLP:journals/tcyb/ManwaniS13} have analyzed the noise tolerance properties of 
 risk minimization under many of the standard loss functions. 
It is shown that risk minimization with $0-1$ loss function is tolerant to uniform noise and 
also to non-uniform noise if the risk of 
optimal classifier under noise-free data is zero \citep{DBLP:journals/tcyb/ManwaniS13}. 
No other loss function is shown to be noise tolerant in this paper (except for square loss under uniform noise). 
It is also shown, through counter-examples, that risk minimization with many of the standard convex loss functions 
(e.g., hinge loss, logistic loss and exponential loss) does not have noise tolerance property  
even under uniform noise \citep{DBLP:journals/tcyb/ManwaniS13}. This paper does not consider the case of 
class-conditional noise. 
A provably correct algorithm  to learn linear classifiers based on risk minimization under 0-1 loss 
is presented in \citep{DBLP:journals/tsmc/SastryNM10}.
This algorithm uses the continuous action-set learning automata~(CALA)~\citep{Thathachar:2003:NLA:1197439}

In this paper we build on and generalize the results presented in  \citet{DBLP:journals/tcyb/ManwaniS13}. 
The main contributions of the paper are the following. 
We provide a sufficient condition on any loss functions such that the risk minimization with that loss function 
becomes noise tolerant under uniform label noise. 
This is a generalization of the main theoretical result in \citet{DBLP:journals/tcyb/ManwaniS13}.
We observe that the $0-1$ loss satisfies this sufficiency condition. We show that ramp loss \citep{DBLP:journals/ior/Brooks11} (which 
is empirically found to be robust in learning from noisy data \citep{Wu07robusttruncated}) 
 and sigmoid loss (which can be viewed as a continuous but non-convex approximation 
of $0-1$ loss) and probit loss \citep{Zheng:2012} also satisfy this sufficiency condition.
We also show that our condition on the loss function along with the assumption that Bayes risk 
(under noise-free distribution) is zero, is sufficient to make risk minimization tolerant to non-uniform noise under 
suitable choice of a parameter in the loss function. We also provide a sufficient condition for robustness to 
class conditional noise. This result generalizes the result presented in  \citet{NIPS2013_5073}. 

In general it is hard to minimize risk under $0-1$ loss.  
Here we investigate approximation of $0-1$ loss function with a differentiable function without losing 
the noise-tolerance property. We show that we can use sigmoid and ramp losses (with some extra 
conditions if we need to tackle nonuniform label noise) for the approximation. We  
investigate standard descent algorithm for minimizing risk under sigmoid and ramp loss.  
Ramp loss can be written as difference of two convex functions~\citep{Wu07robusttruncated}. We make use of this to have an 
efficient algorithm to learn  nonlinear classifiers (through a kernel trick) by minimizing risk under ramp loss.  
We  present extensive empirical investigations to illustrate the noise tolerance properties of 
our risk minimization strategies and compare it against the performance of SVM. 
Among the classifier learning methodologies that can be viewed as risk minimization, Bayes (or Naive Bayes) and SVM 
are the most popular ones. Bayes classifier minimizes risk under $0-1$ loss. Hence we compare performance of risk minimization under 
$0-1$ loss and the other loss functions that satisfy our condition with that of SVM. 

\section{Problem Statement}
\label{sec:problem}

In this paper, our focus is on binary classification. In this section we introduce our notation and 
formally define our notion of noise tolerance of a learning algorithm. 
Here we consider only the 2-class problem.

\subsection{Risk Minimization}
\label{sec:risk-minimization}

We first provide a brief overview of risk minimization for the sake of completeness. More details 
on this can be found in~\citep{Hau1992,DGL1996}. 

Let $\X \subset \R^d$ be the feature space from which
the examples are drawn and let $\Y=\{1,-1\}$ be the class labels. We use $C_+$ and $C_-$ to denote the two classes.
In a typical classifier learning problem, we are given training data,
 $S=\{(\xx_1,y_{\xx_1}),(\xx_2,y_{\xx_2}),\ldots, (\xx_N,y_{\xx_N})\} \in (\X \times \Y)^N$, 
drawn according to an unknown distribution, $\mathcal{D}$,  over $\X \times \Y$. The task is to learn a classifier which 
can predict the class label of a new feature vector. We will represent a classifier as 
$h(\xx) = \mbox{sign}(f(\xx))$ where $f: \X \rightarrow \R$ is a real-valued function defined over the 
feature space. The function $f$ is called a discriminant function though often $f$ is also referred to as the classifier. 
We would use the notation of calling $f$ itself as the classifier though the final prediction of label for a new feature vector is 
given by $\mbox{sign}(f(\xx))$. 

We want to learn a `good' function or classifier from a chosen family of functions, $\F$. 
For example, if we are learning linear classifiers, then $\F =\{W^T\xx + w_0 \; : \; W \in \R^d, \; w_0 \in \R \}$. 
Thus, the family of classifiers of interest here is parameterized by $W, w_0$. 

One way of specifying the goodness of a classifier is through the so called loss function. 
We denote a loss function as $L: \R \times \Y \rightarrow \R^+$. 
The idea is that, given an example $(\xx, y)$, $L(f(\xx), y)$ tells us how well the classifier predicts the label 
on this example. We want to learn a classifier that has, on the average, low loss. Given any loss function, $L$, and 
a classifier, $f$, we define the L-risk of $f$ by 
\begin{equation}
R_L(f) = E [ L(f(\xx), y)]
\end{equation}
where the $E$ denotes expectation with respect to the distribution, $\mathcal{D}$, with which the training examples are drawn. 

Now the objective is to learn a classifier, $f$, that has minimum risk. Such a strategy for learning classifiers 
is called risk minimization. 

As an example, consider the $0-1$ loss function defined by 
\begin{eqnarray}
L_{0-1}(f(\xx), y) & = & 1 \ \mbox{~if~~} yf(\xx) \leq 0 \nonumber \\
                   & = & 0 \ \mbox{~otherwise} \label{0-1loss}
\end{eqnarray}

It is easy to see that the risk under $0-1$ loss of any $f$ is the probability that the classifier $f$ 
misclassifies an example. The Bayes classifier is the minimizer of risk under $0-1$ loss. 

Normally, when one refers to risk of a classifier it is always considered to be under the $0-1$ loss function. Hence, 
here we called the risk under any general loss function as L-risk. This notation is consistent with 
the so called $\phi$-risk used in~\citet{bjm-ccrb-05}. Whenever the specific loss function under consideration is 
clear from context, we simply say risk instead of L-risk.  

Many standard methods of learning classifiers can be viewed as risk minimization with a suitable loss function. 
As noted above, Bayes classifier is same as minimizing risk under $0-1$ loss. Learning a feed-forward neural network based 
classifier can be viewed as risk minimization under squared error loss. (This loss function is defined by $L(a,b) = (a - b)^2$). 
We would mention a few more loss functions later in this paper.

In general, minimizing risk is not feasible because we normally do not have knowledge of the distribution $\mathcal{D}$. 
So, one often approximates the expectation by sample average over the {\em iid} training data and hence one minimizes 
the so called empirical risk given by
\[ \hat{R}_L(f) = \frac{1}{n} \sum_{i=1}^{n}\;  L(f(\xx_i), y_i). \]
If we have sufficient number of training examples (depending on the complexity of the family of classifiers, $\F$), 
then the minimizer of empirical risk would be a good approximation to the minimizer of true risk~\citep{DGL1996}. 
In this paper, all our theoretical results are proved for (true) risk minimization though we briefly comment on 
their relevance to empirical risk minimization.

\subsection{Noise Tolerance}

In this section we formalize our notion of noise tolerance of risk minimization under any loss function. 

Let $S=\{(\xx_1,y_{\xx_1}),(\xx_2,y_{\xx_2}),\ldots, (\xx_N,y_{\xx_N})\} \in (\X \times \Y)^N$ be the (unobservable) 
noise free data, drawn {\em iid} according to a fixed but unknown distribution $\mathcal{D}$ over $\X \times \Y$.
The noisy training data given to learner is $S_{\eta} = \{(\xx_i,\hat{y}_{\xx_i}),i=1,\cdots,N\}$, 
where $\hat{y}_{\xx_i}=y_{\xx_i}$ with probability $(1-\eta_{\xx_i})$ and $\hat{y}_{\xx_i}=-y_{\xx_i}$ with 
probability $\eta_{\xx_i}$. {\em Note that our notation shows that the probability that the label of an example is incorrect 
may be a function of the feature vector of that example.} 
 In general, for a feature vector $\xx$, its correct label (that is, label under distribution $\mathcal{D}$) is denoted as $y_{\xx}$ while the 
noise corrupted label is denoted by $\hat{y}_{\xx}$. We use $\mathcal{D}_{\eta}$ to denote the 
joint probability distribution of $\xx$ and $\hat{y}_{\xx}$. 

We say that the noise is {\em uniform} if $\eta_{\xx}=\eta, \; \forall \xx$. 
Noise is said to be {\em class conditional} if $\eta_{\xx}=\eta_1, \; \forall \xx \in C_+$ 
and $\eta_{\xx}=\eta_2,\; \forall \xx \in C_-$. In general, when noise rate $\eta_{\xx}$ is a function of $\xx$, 
it is termed as {\em non-uniform} noise.

Recall that a loss function is $L: \R \times \Y \rightarrow \R^+$ and in a general risk minimization method,   
we learn a real-valued function $f:\X \rightarrow \R$ by minimizing expectation of loss  
over some chosen function class $\F$. 
For any classifier $f$, the L-risk under  noise-free case is
\[R_L(f)=E_{\mathcal{D}}[L(f(\xx),y_{\xx})]\]

Subscript $\mathcal{D}$ denotes that the expectation is with respect to the distribution $\mathcal{D}$. 
Let $f^*$ be the global minimizer of $R_L(f)$. 

When there is label noise in the data, the data is essentially drawn according to distribution  $\mathcal{D}_{\eta}$. 
 The L-risk of any classifier $f$ under noisy data is
\[R_L^\eta(f)=E_{\mathcal{D}_{\eta}}[L(f(\xx),\hat{y}_{\xx})]\]
Here the expectation is with respect to the joint distribution $\mathcal{D}_{\eta}$ 
which includes averaging over noisy labels also. Let $f^*_{\eta}$ be the global minimizer of risk in the noisy case.
(Note that both $f^*$ and $f^*_{\eta}$ depend on $L$ though our notation does not explicitly show it).

Risk minimization under a given loss function is said to be 
noise tolerant if the $f^*_{\eta}$ has the same probability of misclassification as
that of $f^*$  on the noise free data.
This can be stated more formally as follows~\citep{DBLP:journals/tcyb/ManwaniS13}.

\begin{defn}
Risk minimization under loss function $L$, is said to be {\em noise-tolerant} if
\[P_{\mathcal{D}}[\mbox{sign}(f^*(\xx))=y_{\xx}]=P_{\mathcal{D}}
[\mbox{sign}(f^*_{\eta}(\xx))=y_{\xx}]\].
\end{defn}

When the above is satisfied we also say that the loss function $L$ is noise-tolerant. Note that a loss function can be 
noise tolerant even if the two functions $f^*$ and $f^*_{\eta}$ are different, if both of them have the same classification 
accuracy under the distribution ${\mathcal{D}}$. 
Given a loss function, our goal is to identify, $f^*$ which is a global minimizer of L-risk under the noise-free case. 
If the loss function is noise 
tolerant, then minimizing L-risk with the noisy data would also result in learning $f^*$.


\section{Sufficient Conditions for Noise Tolerance}
\label{sec:theory}

In this section we formally state and prove our theoretical results on noise tolerant risk minimization. 
We start with Theorem~\ref{thm1}, where we provide a sufficient condition for 
a loss function to be noise tolerant under uniform and non-uniform noise.

\begin{thm}\label{thm1}
Let $\eta_{\xx} < .5, \forall {\xx}$. Also, let the loss function $L$ satisfy 
$L(f({\xx}), 1)+L(f({\xx}),-1)=K, \; \forall {\xx}, \; \forall f$ and  for some positive constant $K$. 
Then risk minimization under loss function $L$ becomes noise tolerant under uniform noise. 
If, in addition, $R_L(f^*)=0$, then $L$ is noise tolerant under non-uniform noise also.
\end{thm}
\begin{pf}\hfill 

\begin{itemize}
\item {\bf Uniform Noise: } For any $f$, we have
\[R_L(f) = E_{\mathcal{D}}[L(f(\xx),y_{\xx})] = \int_{\mathcal{X}} L(f({\xx}),y_{\xx})dp({\xx})\].
Under uniform noise, we have $\eta_{\xx} = \eta, \; \forall \xx$. Hence, the L-risk under noisy case for any $f$ is 
\begin{flalign*}
& R_L^\eta(f)=(1-\eta)\int_{\mathcal{X}} L(f({\xx}),y_{\xx})dp({\xx})+ \eta \int_{\mathcal{X}} L(f({\xx}), -y_{\xx})dp({\xx}) \\
&= (1-\eta)\int_{\mathcal{X}} L(f({\xx}),y_{\xx})dp({\xx})+ \eta \int_{\mathcal{X}} (K-L(f({\xx}),y_{\xx}))dp({\xx}) \\
&=R_L(f)(1-2\eta) + K\eta
\end{flalign*}
Hence, $R_L^\eta(f^*)-R_L^\eta(f)=(1-2\eta)(R_L(f^*)-R_L(f)), \; \forall f$. Since $f^*$ is global minimizer of $R_L$, and since 
we assumed $\eta < 0.5$, we get $ R_L^\eta(f^*)-R_L^\eta(f) \leq 0, \forall f$.
Thus $f^* $ is also the global minimizer of $R_L^{\eta}$. 
This completes proof of noise tolerance under uniform noise.
\item {\bf Non-uniform Noise: } Recall that under non-uniform noise, the probability with which a feature vector $\xx$ has 
wrong label is given by $\eta_{\xx}$. Hence, the L-risk under the noisy case for any $f$ is,
\begin{flalign*}
\nonumber &R_L^\eta(f)=\int_{\mathcal{X}} \Big{[}(1- \eta_{\xx}) L(f({\xx}),y_{\xx}) + \eta_{\xx} L(f({\xx}), -y_{\xx})\Big{]}dp({\xx}) \\
\nonumber &=\int_{\mathcal{X}} \Big{[}(1- \eta_{\xx}) L(f({\xx}),y_{\xx})+ \eta_{\xx} \left(K - L(f({\xx}), y_{\xx})\right)\Big{]}dp({\xx}) \\
\nonumber &= \int_{\mathcal{X}}(1- 2\eta_{\xx})L(f({\xx}),y_{\xx})dp({\xx}) + K\int_{\mathcal{X}} \eta_{\xx} dp({\xx})
\end{flalign*}
Hence,

\begin{flalign}
 \nonumber & R_L^\eta(f^*)-R_L^\eta(f)=\int_{\X} (1-2\eta_{\xx}) L(f^*({\xx}),y_{\xx})dp(\xx)\\
&-\int_{\X} (1-2\eta_{\xx})L(f({\xx}),y_{\xx}) dp(\xx)
\label{eq:forbdd}
\end{flalign}
Under our assumption, $R_L(f^*) = \int_{\mathcal{X}} L(f^*({\xx}),y_{\xx})dp({\xx}) = 0$. Since the loss function is 
non-negative, this implies 
 $L(f^*({\xx}),y_{\xx})=0 \; \forall  {\xx}$. Since we assumed $\eta_{\xx} < 0.5, \; \forall \xx$, we have 
$(1-2\eta_{\xx}) \geq 0$. Thus we get  $R_L^\eta(f^*)-R_L^\eta(f) \leq 0, \; \forall f$.
Thus $f^*$ is also global minimizer of risk under non-uniform noise. This proves noise tolerance under non-uniform noise.
\end{itemize}
\end{pf}

The condition on loss function that we assumed in the theorem above is a kind of symmetry condition:
\[ L(f({\xx}), 1)+L(f({\xx}),-1)=K, \; \: \forall {\xx}, \; \forall f.\]
Note that the above condition also implies that the loss function is bounded. Theorem~\ref{thm1} shows that risk minimization under a 
loss function is noise tolerant under uniform noise if the loss function satisfies the above condition. 
For noise tolerance under non-uniform noise, in addition to the above symmetry condition on the loss function, we need $R_L(f^*)=0$. 
In \citet{DBLP:journals/tcyb/ManwaniS13}, this result is proved only for the $0-1$ loss and thus the above theorem is a generalization 
of the main result in that paper. 

Recall that the $0-1$ loss function is given by $L_{0-1}(f(\xx), y_{\xx}) =1$ if $y_{\xx} f(\xx) \leq 0$ and $L_{0-1}(f(\xx), y_{\xx}) =0$ otherwise. 
As is easy to see, the 
$0-1$ loss function satisfies the above symmetry condition with $K=1$. Hence the $0-1$ loss is noise-tolerant under uniform noise. 
None of the standard convex loss functions (such as hinge loss used in SVM or exponential loss used in AdaBoost) 
satisfy the symmetry condition. It is shown in \citet{DBLP:journals/tcyb/ManwaniS13}, through counter-examples,
 that none of them are robust to uniform noise.

\begin{rmk}
For $0-1$ loss to be noise-tolerant under non-uniform noise, 
we need the global minimum of risk under $0-1$ loss to be zero, in the noise-free case. 
This means that, under the noise-free distribution ${\mathcal{D}}$, the classes are 
separable (by a classifier in the family of classifiers  over which we are minimizing the risk). We note that this condition may not 
be as restrictive as it may appear at first sight. This separability is under the noise-free distribution which is, so to say, 
unobservable. For example, consider training data  generated  by sampling from two class conditional densities whose 
supports overlap. We can think of the noise-free data as the one obtained by classifying the data using a Bayes optimal classifier. 
Then the data would be separable under noise-free distribution. The labels in the actual training data could be thought of 
as obtained from this ideal separable data by independent  noise-corruption of the original labels. Then   
 the probability of a label being wrong would  be a function of the feature vector and thus result in 
non-uniform label noise.   
\end{rmk}

If the global minimum of L-risk,  $R_L(f^*)$, is small but non-zero, 
 then we can show that risk minimization under a loss function 
satisfying our symmetry condition would be approximately noise tolerant. Essentially, we can show that 
$R_L(f^*_{\eta})$ can be bounded by $\rho R_L(f^*)$ where $\rho$ is a constant which increases with increasing 
noise rate and would go to infinity as the maximum noise rate approaches 0.5. We derive this bound below. 

Suppose $R_L(f^\ast)= \int_{\X} L(f^*(\xx),y_{\xx}) dp(\xx) = \epsilon$. That is, the global minimum of L-risk under noise-free 
distribution is $\epsilon > 0$. Since $f^\ast_\eta$ is the global minimizer of $R_L^{\eta}$, 
$R_L^\eta(f^\ast)-R_L^\eta(f^\ast_\eta) \geq 0$. From equation~(\ref{eq:forbdd}), we have  
\[
\int_{\X} (1-2\eta_{\xx})\Big(L(f^*(\xx),y_{\xx})-L(f^\ast_\eta(\xx),y_{\xx})\Big)dp(\xx)\geq 0.
\]
This implies 
\[
\int_{\X} (1-2\eta_{\xx}) L(f^\ast_\eta(\xx),y_{\xx})dp(\xx)\leq  \int_{\X} (1-2\eta_{\xx}) L(f^*(\xx),y_{\xx}) dp(\xx) \leq \epsilon
\]
where we used $R_L(f^*)=\epsilon$ and $0 < (1-2\eta_{\xx}) \leq 1$. Let $\eta_{max} = \max_{\xx \in \X} \eta_{\xx}$. Then we have $(1-2\eta_{max}) \int_{\X}  L(f^\ast_\eta(\xx),y_{\xx})dp(\xx) \leq  \epsilon$. Which implies,\[R_L(f^\ast_\eta)\leq  \frac{\epsilon}{1-2\eta_{max}}\].

This shows that if $R_L(f^*)$ is small then $R_L(f^*_{\eta})$ is also small. (Note that 
$f^*_{\eta}$ is what we learn by minimizing risk under the noisy distribution). 
For example, if we have maximum nonuniform noise rate 40\% , then $R_L(f^\ast_\eta)\leq 5\epsilon$.

\begin{rmk}
Our Theorem~\ref{thm1} shows that risk minimization under $0-1$ loss function is tolerant to uniform noise and also to 
non-uniform noise if global minimum of risk is zero. As has been mentioned earlier, the Bayes classifier minimizes risk 
under $0-1$ loss. Hence our result shows that Bayes classifier has good noise tolerance property. 
We can obtain (a good approximation of) Bayes classifier by minimizing risk under $0-1$ loss over an appropriate 
class of functions $\F$. 
We can also obtain (a good approximation of) Bayes classifier 
by estimating the class conditional densities from data. For multidimensional feature vectors, a simplification often employed while 
estimating class conditional densities is to assume independence of features and the resulting classifier is termed Naive Bayes 
classifier. In many situations this would be a good approximation to Bayes classifier. In a recent study, Nettleton et al. presented extensive 
empirical investigations on noise robustness of different classifier learning algorithms~\citep{NOF2010}. 
In their study, they considered the top ten 
machine learning algorithms~\citep{YZSHS2007}. They found that the Naive Bayes classifier has the best robustness 
with respect to noise. Theorem~\ref{thm1} proved above provides some theoretical justification for the noise-robustness 
of Naive Bayes classifier. Later, in Section~\ref{sec:experiments} we also present simulation results to 
show that risk minimization under $0-1$ loss has very good robustness to label noise.  
\end{rmk}

\begin{rmk}
As mentioned in Section~\ref{sec:risk-minimization}, in practice one minimizes empirical risk because 
one often does not have the knowledge of class conditional densities. Our theorem, as proved, applies only to (true) risk minimization. 
If we have good number of examples and if the complexity of the class of function $\F$ is not large, then, by the standard 
results on consistency of empirical risk minimization~\citep{DGL1996}, the minimizer of empirical risk under noise free distribution would be 
close to minimizer of true risk under noise-free distribution and similarly for the noisy distribution. Hence, it is reasonable to 
assume that minimizer of empirical risk with noisy samples would be close to minimizer of empirical risk with noise-free samples. 
Also, if we take the expectation integral in the proof of Theorem~\ref{thm1} to be with respect to the empirical distribution given by 
the given set of examples, then the L-risk under noise-free distribution is same as the empirical risk. Then Theorem~\ref{thm1} 
can be interpreted as saying that the minimizer of empirical risk with noise-free samples would be same as the minimizer of 
empirical risk with noisy samples averaged over the label-noise distribution. All this provides a plausibility argument that the 
noise-robustness property proved by Theorem~\ref{thm1} would (approximately) hold even for the case of empirical risk minimization. 
Our empirical results presented in Section~\ref{sec:experiments} also provide evidence for this.  
More work is needed to formally prove such a result to extend the noise-robustness results to empirical risk minimization  
and to derive some bounds on the number of examples needed.  
\end{rmk}

Risk minimization under $0-1$ loss is  hard because it involves optimizing a non-convex and non-smooth objective function. 
One can easily design a smooth loss function ( which can be viewed as a continuous approximation of the $0-1$ loss function) 
 that can satisfy the symmetry condition of Theorem~\ref{thm1}. Hence, one can try 
optimizing risk under such a loss function. 
As we show here, we can use the ramp loss, the sigmoid loss etc. for this. 
However, under such a loss function, it may not be possible to achieve $R_L(f^*)=0$. For example, 
 a sigmoid function value is always strictly positive and hence the 
risk (under such a loss function) of any classifier is strictly greater than zero. 
Thus for other loss functions which can satisfy our symmetry condition, the sufficient condition 
for noise tolerance under non-uniform noise, namely that global minimum of L-risk (under that loss function) is zero, may be 
very restrictive. We address this issue next.

We call the global minimum of risk under $0-1$ loss as Bayes risk. If we assume that Bayes risk under noise-free case 
is zero, then we can show that some of the loss 
functions satisfying our symmetry condition 
can achieve noise tolerance under non-uniform noise also by proper choice of a parameter in the loss function (even if the 
global minimum of L-risk is non-zero).  
We present these results for the sigmoid loss, the ramp loss and the probit loss in the next three subsections. 

\subsection{Sigmoid Loss} Sigmoid loss with parameter $\beta > 0$ is defined as
\begin{equation}
L_{\mbox{sig}}(f({\xx}),y_{\xx})=\frac{1}{1+\exp(\beta f({\xx}) y_{\xx})}
\end{equation} 
If we view the loss as a function of the single variable $f(\xx)y_{\xx}$, then the parameter $\beta$ is proportional 
to the magnitude of the slope of the function at origin.  It is easy to verify that 
\[L_{\mbox{sig}}(f({\xx}),1)+L_{\mbox{sig}}(f({\xx}),-1)=1,\; \forall {\xx},\; \forall f.\]
The following theorem shows that sigmoid loss function is noise tolerant.
\begin{thm}\label{thm3}
 Assume $\eta_{\xx} <.5,\; \forall {\xx}$. Then sigmoid loss is noise tolerant under uniform noise. 
In addition, if Bayes risk under noise-free case is zero, then there exist a constant $\beta_M  < \infty$ 
such that $\forall \beta\geq \beta_M$ the risk minimization under sigmoid loss is tolerant to non-uniform noise.
\end{thm}
\begin{pf}
First part of the theorem follows directly from Theorem~\ref{thm1} because sigmoid loss satisfies the symmetry condition. 
We prove second part below.
For any $f$, the L-risk under the noisy case is given by

\begin{flalign*}
& R_L^\eta(f)=\int_{\mathcal{X}} \Big{[}(1- \eta_{\xx}) L_{\mbox{sig}}(f({\xx}),y_{\xx}) + 
\eta_{\xx} L_{\mbox{sig}}(f({\xx}), -y_{\xx}) \Big{]}dp({\xx}) \\
&=\int_{\mathcal{X}} \Big{[}(1- \eta_{\xx}) L_{\mbox{sig}}(f({\xx}),y_{\xx}) + 
\eta_{\xx} (1 - L_{\mbox{sig}}(f({\xx}), y_{\xx})) \Big{]} dp({\xx}) \\
&=\int_{\mathcal{X}} \eta_{\xx} dp({\xx})+ \int_{\mathcal{X}}(1- 2\eta_{\xx})L_{\mbox{sig}}(f({\xx}),y_{\xx})dp({\xx}) \\
&= \int_{\mathcal{X}} \eta_{\xx} dp({\xx}) + \int_{\mathcal{X}} (1 -2\eta_{\xx}) \frac{1}{1+\exp(\beta f({\xx})y_{\xx})} dp({\xx})
\end{flalign*}
Hence,
\begin{flalign}
\nonumber & R_L^\eta (f^*)-R^\eta(f)\\
&=\int_{\mathcal{X}} (1-2\eta_{\xx})\Big(\frac{1}{1+\exp(\beta f^*({\xx})y_{\xx})} - \frac{1}{1+\exp(\beta f({\xx})y_{\xx})}\Big)dp({\xx}) 
\label{eq:diff}
\end{flalign} 
For establishing noise tolerance under non-uniform noise, 
we need to show that, $R_L^\eta (f^*)-R_L^\eta(f)<0,\; \forall \beta>\beta_M, \; \forall f$. We define three sets $S_1$, $S_2$, $S_3$ where,
$S_1 = \{{\xx}:f({\xx})y_{\xx}<0\}$, $S_2 = \{{\xx}: f^*({\xx})y_{\xx} < f({\xx})y_{\xx}\}$
and $S_3 = \{{\xx}: f^*({\xx})y_{\xx} \geq f({\xx})y_{\xx} \geq 0\}$.

 Since we assumed that Bayes risk (under noise-free case) is $0$, $f^*({\xx})y_{\xx}>0, \; \forall \xx$.
  Note that the three sets above  form a partition of $\mathcal{X}$.  Now we can rewrite equation~(\ref{eq:diff}) as 
\begin{flalign}
\nonumber & R_L^\eta (f^*)-R_L^\eta(f)\\
\nonumber & = \int_{S_1} (1-2\eta_{\xx})\Big(\frac{1}{1+\exp(\beta f^*({\xx})y_{\xx})} - \frac{1}{1+\exp(\beta f({\xx})y_{\xx})}\Big)dp({\xx}) \\
\nonumber & +\int_{S_2}(1-2\eta_{\xx})\Big(\frac{1}{1+\exp(\beta f^*({\xx})y_{\xx})} - \frac{1}{1+\exp(\beta f({\xx})y_{\xx})}\Big)dp({\xx})\\
& +\int_{S_3}(1-2\eta_{\xx})\Big(\frac{1}{1+\exp(\beta f^*({\xx})y_{\xx})} - \frac{1}{1+\exp(\beta f({\xx})y_{\xx})}\Big)dp({\xx})
\label{eq:split}
\end{flalign}
We observe the following.
\begin{itemize}
\item The third term is less than or equal to zero always because, on $S_3$, we have 
$0\leq f({\xx})y_{\xx}\leq f^*({\xx})y_{\xx}$. 
\item The first integral is over $S_1$ where we have $f({\xx})y_{\xx}<0<f^*({\xx})y_{\xx}$. 
Since $(1-2\eta_{\xx})>0$, the integral has negative value for all $\beta$. 
The value of this integral decreases with increasing $\beta$. As $\beta \rightarrow \infty$, the integral 
becomes $-M<0$, where $M=\int_S (1-2\eta_{\xx})dp({\xx})$. We have $M$ strictly greater than zero, 
because if $f$ is not the optimal classifier then $\int_{S_1} dp({\xx}) >0$.
\item The second  integral is over $S_2$, where $0<f^*({\xx})y_{\xx}<f({\xx})y_{\xx}$. 
This integral is always positive and as $\beta\rightarrow \infty $, the limit of the integral is zero.
\end{itemize}
Thus as $\beta \rightarrow \infty$, the limit of the sum of first two terms on the RHS of equation~(\ref{eq:split}) is $-M<0$. 
Hence there exist a $\beta_M$ such that for all $\beta>\beta_M$, the sum of first two integral is negative. 
The third term on the RHS of equation~(\ref{eq:split}) is always non-positive. 
This shows that for all $\beta>\beta_M$, $R^\eta (f^*)-R^\eta(f)<0$ and this completes the proof.
\end{pf}

Theorem~\ref{thm3} shows that if we take a sufficiently large value of the parameter $\beta$, then sigmoid loss is noise tolerant 
under non-uniform noise also. This is so even though the global minimum of risk, in the noise-free case, under sigmoid loss 
is greater than zero. (But we assumed that  the Bayes risk under noise-free 
case is zero). What this means is that we need the loss function (as a function of the variable $f(\xx)y$) 
to be sufficiently steep at origin to well-approximation of $0-1$ loss so as to get noise tolerance.
We also note here that the value of $\beta_M$, which may be problem dependent, can be fixed through cross validation in practice.

\subsection{Ramp Loss}
Ramp loss with a parameter $\beta > 0$ is defined by,
\begin{equation}\label{RampLoss}
L_{\mbox{ramp}}(f({\xx}),y_{\xx})=(1-\beta f({\xx})y_{\xx})_{+}  \; - \; (-1-\beta f({\xx})y_{\xx})_{+}
\end{equation} 
where $(A)_+$ denotes the positive part of $A$ which is given by $ A_+ = 0.5(A + |A|)$. 
The following lemma shows that the ramp loss function satisfies the symmetry property needed in Theorem~\ref{thm1}.
\begin{lem}\label{lemma1}
Ramp Loss described in Eq.~(\ref{RampLoss}) satisfies 
\begin{equation}
\nonumber L_{\mbox{ramp}}(f({\xx}),y_{\xx})+L_{\mbox{ramp}}(f({\xx}),-y_{\xx})=2,\; \forall {\xx},\; \forall f
\end{equation}
\end{lem}
\begin{pf} 
We have 
\begin{flalign*}
 & L_{\mbox{ramp}}(f({\xx}),y_{\xx})+L_{\mbox{ramp}}(f({\xx}),-y_{\xx})  \\
&= (1-\beta y_{\xx}f({\xx}))_{+} - (-1-\beta y_{\xx}f({\xx}))_{+} + (1+\beta y_{\xx}f({\xx}))_{+}\\
& \quad - (-1+\beta y_{\xx}f({\xx}))_{+}  \\
&= \frac{1}{2}\big{[}(1-\beta y_{\xx}f({\xx}))-|1-\beta y_{\xx}f({\xx})|\big{]}-\frac{1}{2}\big{[}(-1-\beta y_{\xx}f({\xx}))\\
& \quad + |1+\beta y_{\xx}f({\xx})|\big{]} +\frac{1}{2}\big{[}(1+\beta y_{\xx}f({\xx}))-|1+\beta y_{\xx}f({\xx})|\big{]}\\
& \quad -\frac{1}{2}\big{[}(-1+\beta y_{\xx}f({\xx})) + |1-\beta y_{\xx}f({\xx})|\big{]} \\
&= 2  
\end{flalign*}
which completes the proof.
\end{pf}

The above lemma shows that the ramp loss satisfies our symmetry condition and hence, by Theorem~\ref{thm1}, is noise-tolerant to 
uniform noise. 
It has been empirically observed that ramp loss is more robust to noise than 
SVM \citep{Wu07robusttruncated,DBLP:conf/aaai/XuCS06,DBLP:journals/ior/Brooks11}. Our results provide a theoretical justification for it.

The following theorem shows that ramp loss can be noise-tolerant to non-uniform noise also if $\beta$ is sufficiently high. 
\begin{thm}\label{thm4} Assume $\eta_{\xx} <.5, \forall {\xx}$. Then the ramp loss is noise tolerant under uniform noise. 
Also, if Bayes risk under noise-free case is zero, there exist a constant $\beta_M < \infty$ such that 
$\forall \beta\geq \beta_M$ the risk minimization under ramp loss is tolerant to non-uniform noise.
\end{thm}
\begin{pf} Lemma~\ref{lemma1} shows that the ramp loss satisfies the symmetry property. 
Thus, Theorem~\ref{thm1} directly implies that ramp loss is noise tolerant under uniform noise. 
Proof of noise tolerance under non-uniform noise is similar to proof of Theorem~\ref{thm3} 
 and it follows from the same decomposition of feature space. We omit the details. 
\end{pf}

\subsection{Probit Loss}
Probit loss \citep{Zheng:2012,conf/nips/McAllesterK11} with a parameter $\beta > 0 $ is defined by,
\begin{equation}\label{ProbitLoss}
L_{\mbox{probit}}(f({\xx}),y_{\xx})=1-\Phi(\beta f({\xx})y_{\xx})
\end{equation}
where $\Phi$ is cumulative distribution function (CDF) of standard Normal distribution.
\begin{lem}\label{lemma2}
Probit Loss described in Eq.~(\ref{ProbitLoss}) satisfies 
\begin{equation}
\nonumber L_{\mbox{probit}}(f({\xx}),y_{\xx})+L_{\mbox{probit}}(f({\xx}),-y_{\xx})=1,\; \forall {\xx},\; \forall f
\end{equation}
\end{lem}
\begin{pf} 
\begin{flalign*}
 &L_{\mbox{probit}}(f({\xx}),y_{\xx})+L_{\mbox{probit}}(f({\xx}),-y_{\xx})\\
& = 1-\Phi(\beta f({\xx})y_{\xx})+1-\Phi(-\beta f({\xx})y_{\xx}) = 1  
\end{flalign*} 
because $\Phi(-z) = 1 - \Phi(z), \; \forall z \in \R$.
Hence $L_{\mbox{probit}}$ satisfies the symmetry property.
\end{pf}

\begin{thm}\label{thm5} Assume $\eta_{\xx} <.5, \forall {\xx}$. 
Then probit loss is noise tolerant under uniform noise. 
Also, if Bayes risk under noise-free case is zero, there exists a constant $\beta_M < \infty$ 
such that $\forall \beta\geq \beta_M$ the risk minimization under probit loss is tolerant to non-uniform noise.
\end{thm}
\begin{pf} Lemma~\ref{lemma2} shows that the probit loss satisfies the symmetry property. 
Thus, Theorem~\ref{thm1} directly implies that probit loss is noise tolerant under uniform noise. 
Proof of noise tolerance under non-uniform noise is similar to proof of Theorem~\ref{thm3}
 and it follows from the same decomposition of feature space. We omit the details.
\end{pf}

\subsection{Class-conditional Noise}

So far, we have considered only the cases of uniform and non-uniform noise. A special case of non-uniform noise is 
class conditional noise where noise rate is same for all feature vectors from one class. This is an interesting 
special case of label noise~\citep{stempfel2007learning,conf/colt/ScottBH13,NIPS2013_5073}. 
In the results proved so far, we need Bayes risk under noise-free case to be zero for a loss function to be 
tolerant to non-uniform noise. Since class conditional noise is a very special case of non-uniform noise, an 
interesting question is to ask whether this condition can be relaxed. 

Under class conditional noise we have $\eta_{\xx}=\eta_1,\;\forall {\xx}\in C_+ \;\&\; \eta_{\xx}=\eta_2, \forall {\xx}\in C_-$. 
Suppose we know $\eta_1$ and $\eta_2$. Note that this does not make the problem trivial because we still do not know which are 
the examples with wrong labels. It may be possible to estimate the noise rates from the noisy training data using, e.g., the method 
in~\citet{conf/colt/ScottBH13}. In such a situation, we can ask how to make risk minimization noise tolerant. 
Suppose we have a loss function $L$ that satisfies our symmetry condition. The following theorem shows how we can  learn global minimizer 
of L-risk  under the noise-free case given access only to data corrupted with class conditional label noise. 


\begin{thm}\label{thm2}
 Assume $\eta_{\xx}=\eta_1,\;\forall {\xx}\in C_+ \;\&\; \eta_{\xx}=\eta_2, \forall {\xx}\in C_-$, and $\eta_1+\eta_2 <1$. 
Assume loss function $L(.,.)$ satisfies, for some positive constant $K$, $L(f({\xx}),1)+L(f({\xx}), -1)=K, \; \forall \xx, \; \forall f$. 
We define loss function $l(.,.)$ as $l(f({\xx}),1)=L(f({\xx}),1)$ \& $l(f({\xx}),-1)=kL(f({\xx}),-1) $ 
where $k=\frac{1-\eta_1+\eta_2}{1-\eta_2+\eta_1}$. Then minimizer of risk with  loss function $l(.,.)$ under 
class conditional noise is same as minimizer of risk with loss $L(.,.)$ under noise free data.
\end{thm}
\begin{pf}
For any $f$, under no noise, we have,
\begin{eqnarray}
R(f)&=&\int_{\xx} L(f({\xx}),y_{\xx})dp({\xx}) \nonumber
\end{eqnarray}
Under class conditional noise, we use the loss function $l(.,.)$, and hence the risk under noisy case is 
\begin{flalign*}
&R^\eta(f)=\int_{{\xx}\in C_+}\Big{[}(1-\eta_1)l(f({\xx}),1) + \eta_1 l(f({\xx}),-1)\Big{]}dp({\xx}) \\
& + \int_{{\xx}\in C_-}\Big{[}(1-\eta_2)l(f({\xx}),-1) +\eta_2 l(f({\xx}),1)\Big{]}dp({\xx}) \nonumber \\
&= \int_{{\xx}\in C_+}\Big{[}(1-\eta_1)L(f({\xx}),1) + \eta_1 k L(f({\xx}), -1)\Big{]}dp({\xx}) \nonumber \\
& + \int_{{\xx}\in C_-}\Big{[}(1-\eta_2)k L(f({\xx}), -1) +\eta_2 L(f({\xx}),1) \Big{]}dp({\xx}) \nonumber \\
&= \int_{{\xx}\in C_+}\Big{[}(1-\eta_1)L(f({\xx}),1) + \eta_1 k (K - L(f({\xx}),1))\Big{]} dp({\xx}) \nonumber \\
& + \int_{{\xx}\in C_-}\Big{[}(1-\eta_2)k L(f({\xx}), -1) +\eta_2(K - L(f({\xx}), -1)) \Big{]}dp({\xx}) \nonumber 
\end{flalign*}
It is easy to see that, with the value of $k$ given in the theorem statement, we have 
$(1 - \eta_1) - \eta_1 k = (1 - \eta_2)k - \eta_2$. Using this in the above, we get
\begin{flalign*}
& R^{\eta}(f) = \frac{1-\eta_1-\eta_2}{1-\eta_2+\eta_1}\Big{[}\int_{{\xx}\in C_+}L(f({\xx}),1)dp({\xx})\\
&+\int_{{\xx}\in C_-}L(f({\xx}), -1)dp({\xx})\Big{]}+\mbox{const}\nonumber \\
&= \frac{1-\eta_1-\eta_2}{1-\eta_2+\eta_1}R(f)+\mbox{const}
\end{flalign*}
Hence,
\begin{equation}
\nonumber R^\eta(f^*)-R^\eta(f)=\frac{1-\eta_1-\eta_2}{1-\eta_2+\eta_1} [R(f^*)-R(f)].
\end{equation}
As $(1-\eta_1-\eta_2)>0$ and $(1-\eta_2+\eta_1)>0$, we have $R^\eta(f^*)-R^\eta(f)\leq 0, \; \forall f$. 
Thus $f^*$, which is global minimizer of risk with loss function $L$ under noise-free data 
 is also the global minimizer of risk under class conditional noise with loss function $l(.,.)$.
\end{pf}

The above theorem allows us to construct a new loss function $l$  given the loss function $L$ (and the noise rates) so that 
minimizing risk under the noisy case with loss $l$ would result in learning minimizer of risk with $L$ under noise-free data. 

The special case of this theorem when $L$ is the $0-1$ loss function is proved in~\citet{NIPS2013_5073}. Hence, 
Theorem~\ref{thm2} is a generalization of their result to any loss function that satisfies our symmetry condition 
(such as sigmoid loss or ramp loss).

\section{Experiments}
\label{sec:experiments}
In this section, we present some empirical results on both synthetic and real data sets to illustrate the 
noise tolerance properties of different loss functions. Our theoretical results 
have shown that $0-1$ loss, sigmoid loss and ramp loss are all  noise tolerant. 
We compare performances of risk minimization with these noise tolerant losses with SVM which is hinge loss based risk minimization  
approach. Square loss has also been shown to be noise tolerant under uniform label noise \citep{DBLP:journals/tcyb/ManwaniS13}. 
Hence we also compare with square loss. 
The experimental results are shown on $5$ synthetic datasets and $5$ real world datasets from UCI ML repository~\citep{Bache+Lichman:2013}. 

\subsection{Dataset Description}
We used 5 synthetic problems of 2-class classification. Among these, 4 problems are linear and 1 is non-linear.
All synthetic problems have separable classes under noise-free case. We consider both two dimensional data (so that we can 
geometrically see the performance) as well as higher dimensional data (with dimension $d=50$).  
Below, we describe each of the synthetic problems by describing how the labeled training data is generated 
 under noise-free case. 
We add label noise as needed to generate noisy training sets. 
 In the description below we denote the uniform density function with support set $A$ by $\U(A)$. 

\begin{enumerate}
\item {\bf Synthetic Dataset~1 : Uniform Distribution}
In $\R^{20}$, we sample $3000$ {\em iid} points from $\U([-1\;\;1]^{20})$. 
We label these samples using the following separating hyperplane.
\begin{eqnarray}
\nonumber \ww_1 = [\; \o^{10}\;\;-\o^{10}\;],\;\;\;\; b_1=0  \nonumber
\end{eqnarray}
where $\o^{10}$ is a 10-dimensional vector of 1's.

\item {\bf Synthetic Dataset~2 : Asymmetry and Non-uniformity}
Let $f_1$ and $f_2$ be two mixture density functions in $\R^2$ defined as follows 
\begin{flalign*}
&f_1(\xx)= 0.45\;\U([-1,0]\times [-1,1])+0.5\;\U([-4,-3]\times [0,1]) \nonumber \\
& +0.05\;\U([-10,0]\times [-5,5]) \nonumber \\
& f_2(\xx)= 0.45\;\U([0,1]\times [-1,1])+0.5\;\U([9,10]\times [-1,0]) \nonumber \\
& +0.05\;\U([0,10]\times [-5,5]) \nonumber
\end{flalign*}
 We sample 2000 {\em iid} points each from $f_1$ and $f_2$.
We label these points using the following hyperplane
\begin{equation}
\nonumber \ww_2=[1\;\;\;0],\;\;\;\;b_2=0
\end{equation}

\item {\bf Synthetic Dataset~3 : Asymmetry and Imbalance}
Let $f_1$ and $f_2$ be two  density functions in $\R^2$ defined as follows
\begin{eqnarray}
f_1(\xx) &=& \U([-10.1,-0.1]\times [-5,5]), \nonumber \\
f_2(\xx) &=& \U([0.1,1.1]\times [-2.5,2.5]). \nonumber
\end{eqnarray}
We sample $3000$ points independently from $f_1$ and $1000$ points independently from distribution $f_2$.
We label these points using the following hyperplane
\begin{equation}
\nonumber  \ww_3=[1\;\;\;0],\;\;\;\;b_3=0
\end{equation}

\item {\bf Synthetic Dataset~4 : Asymmetry and Imbalance in High Dimension}
Let $f_1$ and $f_2$ be two uniform densities defined in $\R^{50}$ as follows
\begin{eqnarray}
\nonumber f_1 &=& \U([-10.1,-0.1]\times [-2.5,2.5]^{49}), \\
\nonumber f_2 &=& \U([0.1,1.1]\times [-1,1]^{49}).
\end{eqnarray}
We sample $8000$ and $4000$ points independently from $f_1$ and $f_2$ respectively.
We label these points using the following hyperplane.
\begin{eqnarray}
\nonumber \ww_4 = \ee^{50},\;\; b_4=0
\end{eqnarray} 
where $\ee^{50}$ is the standard basis vector in $\R^{50}$ whose first element is 1 and rest of all
are 0.

\item {\bf Synthetic Dataset~5 :  2$\times$2 Checker Board}
Let $f$ be a uniform density defined on $\R^2$ as follows
\begin{eqnarray}
\nonumber f = \U([0,4]\times [0,4])
\end{eqnarray}
We sample 4000 points independently from $f$. We classify these points using $\mbox{sign}(x_1-2)(x_2-2)$,
where $x_1$ and $x_2$ represent the first and the second dimension of $\R^2$.
\end{enumerate}

Apart from the above synthetic data sets we also consider 5 data sets from the UCI ML repository described in Table~\ref{UCI}.
\begin{table}
\begin{center}
\caption{Dataset Used from UCI ML Repository}
\label{UCI}
\begin{tabular}{cccc}
\hline
\hline
Dataset  &  \# Points & Dimension & Class Dist.\\
\hline
\hline
Ionosphere   & 351   & 34 & 225,126 \\
Balance  &  576 &  4 & 288,288   \\
Vote &  435 &  15  & 267,168  \\
Heart & 270  & 13 & 120,150 \\
WBC  & 683 &  10  & 239,444 \\
\hline
\hline
\end{tabular}
\end{center}
\end{table}

\subsection{Experimental Setup} 
We implemented all risk minimization algorithms in MATLAB. 
There is no general purpose algorithm for minimizing empirical risk under $0-1$ loss. 
We use the method based on a team of continuous action-set learning automata (CALA)~\citep{DBLP:journals/tsmc/SastryNM10}. 
It is known that if the step-size parameter, $\lambda$, is sufficiently small, 
CALA-team based algorithm converges to global minimum of risk  in linear classifier case \citep{DBLP:journals/tsmc/SastryNM10}. 
In our simulations, we keep $\lambda=5\times 10^{-5}$. Since this algorithm takes a little long to converge, we show results 
for risk minimization with $0$-$1$ loss only on Synthetic dataset~1 and on Breast Cancer dataset. 

For risk minimization with ramp loss and sigmoid loss for learning linear classifiers, 
we used simple gradient descent with decreasing step size and a momentum term. We use an incremental version; that is 
we keep updating the linear classifier after processing each example and we choose the next example randomly from the training data. 
The gradient descent is run with multiple starts ($3$ times) and we keep the best final value. 
We learn with $\beta = 2, \; 4$ when we have uniform noise and with $\beta=4, \; 8, \; 12$ when we have non-uniform (or class conditional) noise. 
In all cases we  report the results with best $\beta$ value.

We illustrate learning of nonlinear classifiers only with minimizing risk under ramp loss. 
The regularized (empirical) risk under ramp loss can be written as difference of two convex functions. 
This decomposition leads to an efficient minimization algorithm 
 using DC (difference of convex) program \citep{Pham1997,Wu07robusttruncated}. 
DC algorithm for learning a nonlinear classifier
by minimizing regularized risk under ramp loss  is 
explained in \ref{appdx:1}. This is the method (as described in Algorithm~\ref{algo2}) we used to learn nonlinear classifiers. 
 We compared ramp loss based classifier with SVM (based on hinge loss) for nonlinear problems.

To learn SVM classifier, we used LibSVM code \citep{CC01a}. We have run experiments with different values of the SVM parameter, $C$ ($C=10, \; 100, \; 500, \; 1000$) and the results reported are those with best $C$ value.  

In the previous subsection, we explained how the noise-free data is generated for synthetic problems. For the bench mark data sets we take the data as noise free. We then randomly add uniform or non-uniform or class conditional (CC) noise. For 
uniform noise case we vary the noise rate ($\eta$) from $10\%$ to $40\%$. For class conditional noise we used rates of $30\%$ and $10\%$. 
 We incorporate non-uniform noise as follows. 
For every example, the probability of flipping the label is based on which quadrant (with respect to its first two features) 
the example falls in. For  non-uniform noise, the rates in the four quadrants are $35\%,30\%,25\%,20\%$ respectively for all problem.

For each problem, we randomly used $75\%$ for training (within training data, $33\%$ is used for validation) and $25\%$ for test sets.
Then the training data is corrupted with label noise as needed. We determine the accuracy of the learnt classifier
on the test set which is noise-free. In each case, this process of random choice of training and test sets is repeated $10$ times.
We report the average (and standard deviation) of  accuracy of different methods for different noise rates.

\subsection{Simulation Results on Synthetic Problems}
\begin{figure}
\begin{center}
\includegraphics[scale=0.17]{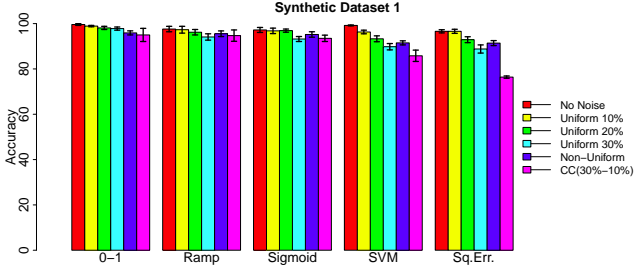}
\caption{Comparison Results on Synthetic Dataset 1}
\label{P1}
\end{center}
\end{figure}

\begin{figure}
 \begin{center}
  \begin{tabular}{cc}
\includegraphics[scale=.3]{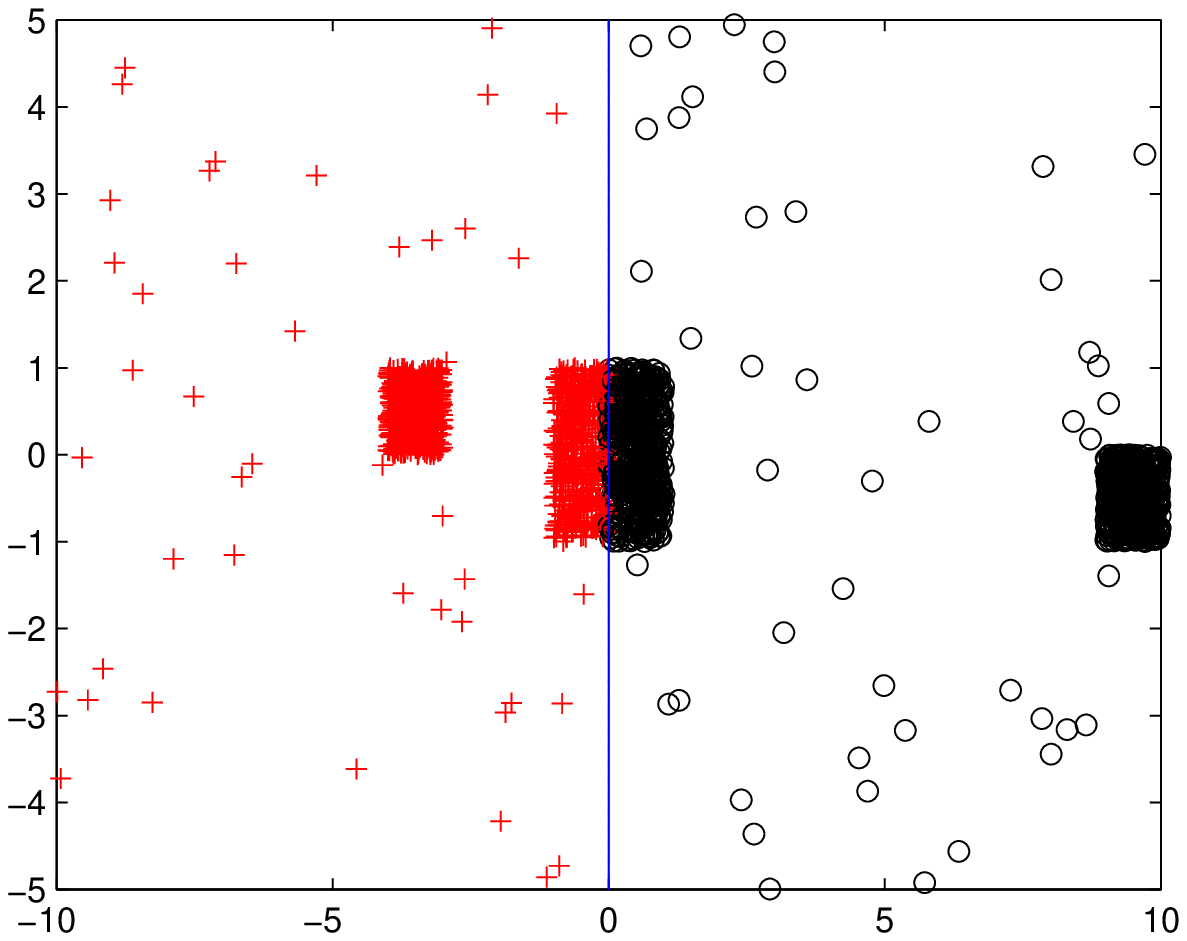}  &  \includegraphics[scale=.3]{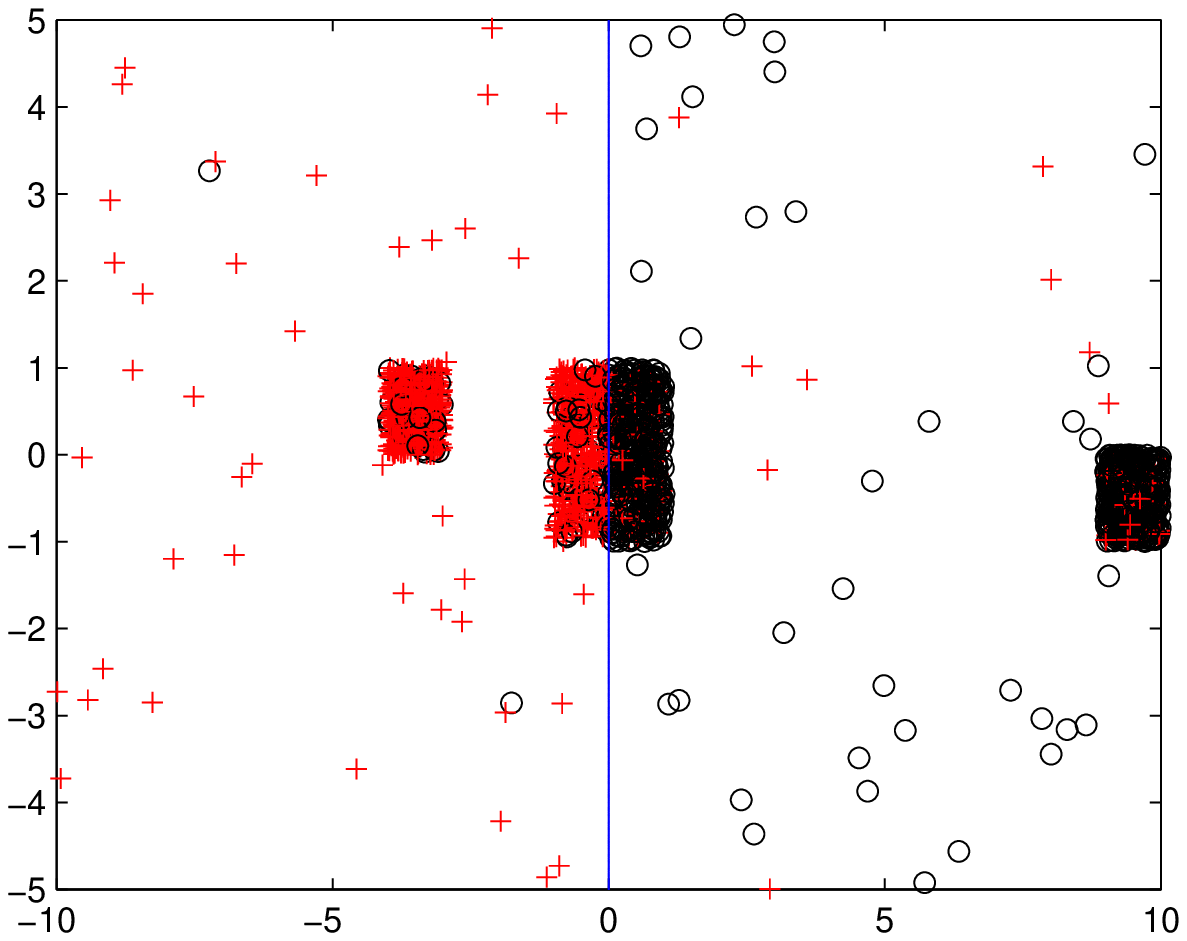}\\
(a) & (b) \\
\includegraphics[scale=.3]{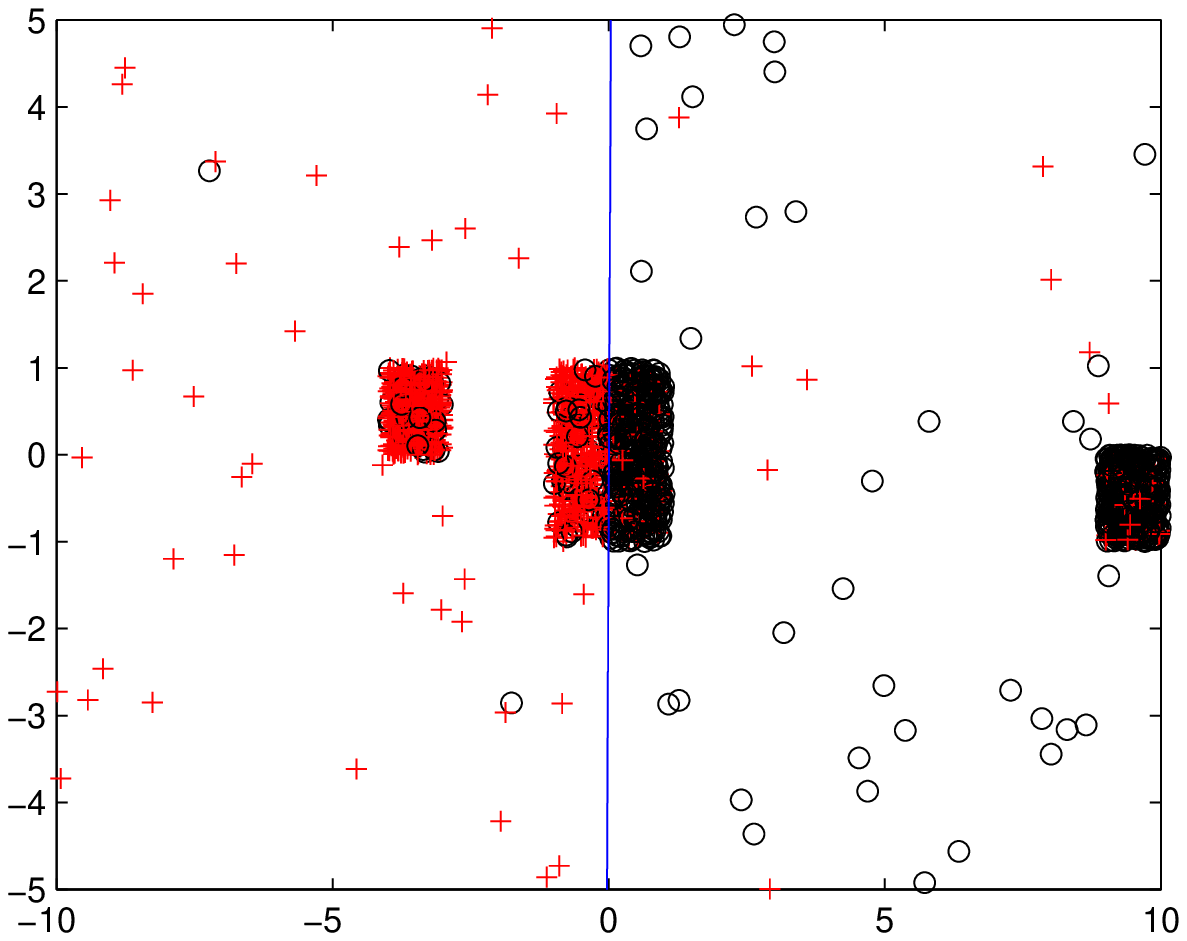}       &  \includegraphics[scale=.3]{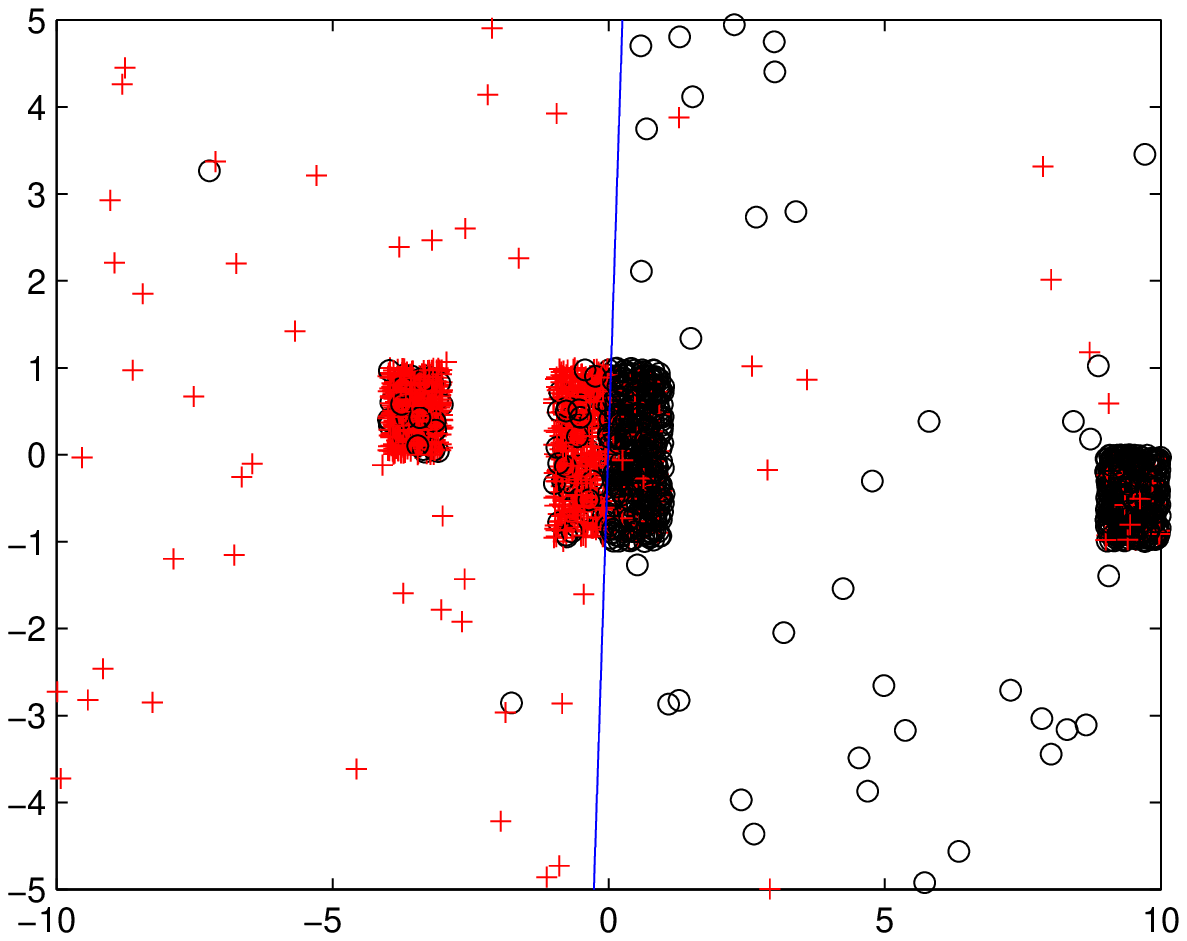} \\
(c)  &  (d) \\
\includegraphics[scale=.3]{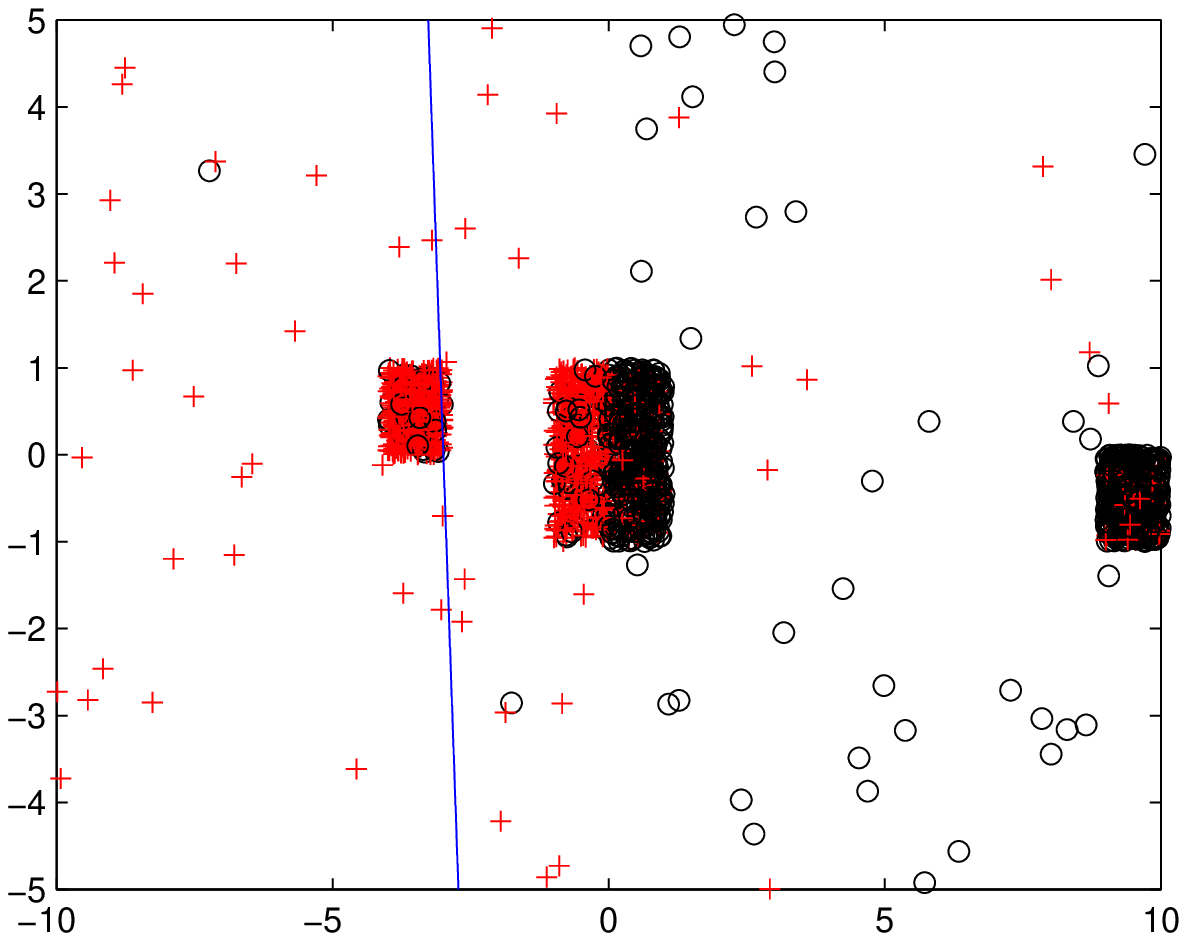}       &  \includegraphics[scale=.3]{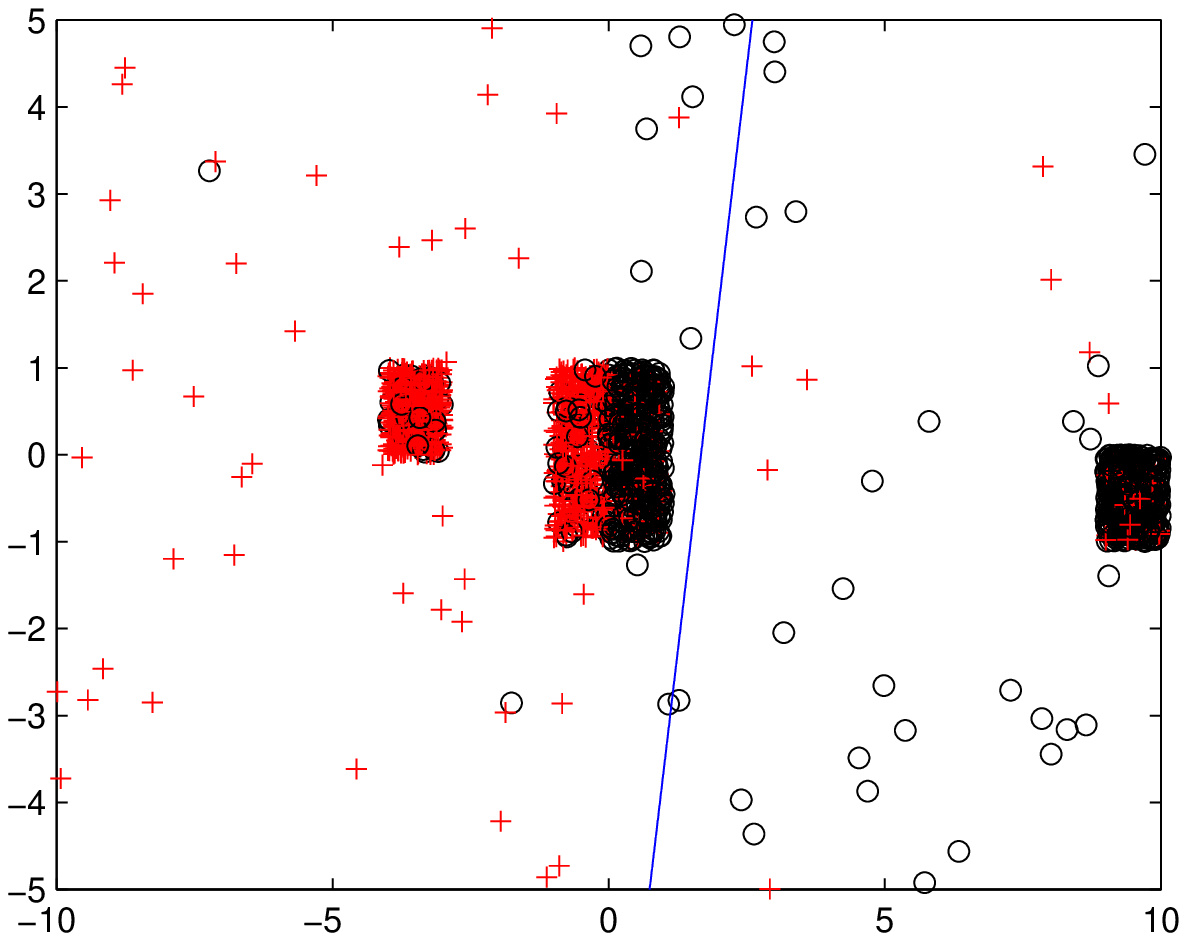}   \\
(e) & (f)
  \end{tabular}
\caption{\footnotesize{Results on Synthetic Dataset 2. (a) the data along with true classifier (Solid line), 
(b) data corrupted with 10\% uniform noise. Linear classifiers learnt by minimizing (c) sigmoid loss (d)ramp loss
(e) hinge loss (linear SVM) (f) Square loss.}}
\label{fig:1d-example6}
 \end{center}
\end{figure}

\begin{figure}
\begin{center}
\includegraphics[scale=0.17]{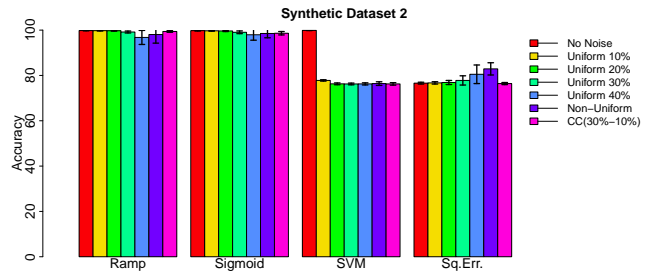}
\caption{Comparison Results on Synthetic Dataset 2}
\label{P2}
\end{center}
\end{figure}
In Synthetic Dataset~1, classes are symmetric with uniform class conditional densities and the examples from the two classes are balanced. As can be seen from Figure~\ref{P1}, accuracy of $0-1$ loss drops to only $97.8\%$, sigmoid loss and ramp loss accuracies drop to $93\%$ but accuracy of SVM drops severely to $89.8\%$. Under non-uniform noise, sigmoid loss, ramp loss, $0-1$ loss perform much better than SVM. Under class conditional noise, SVM's accuracy drops to $86\%$, whereas all the noise-tolerant losses have accuracy around $95\%$.

In Synthetic Dataset 2, we have balanced but asymmetric classes in $\R^2$. In addition to that we 
have nonuniform class conditional densities. Figure~\ref{fig:1d-example6} presents classifiers learnt using sigmoid loss, ramp loss, hinge loss and square error loss on Synthetic Dataset~2 with 10\% uniform label noise. We see that sigmoid loss and ramp loss based risk minimization approaches accurately capture the true classifier.
On the other hand, SVM (hinge loss) and square error based approach fail to learn the true classifier in presence of label noise. As can be seen from Figure~\ref{P2}, even under $10\%$ noise, 
accuracy of SVM drops to $77.8\%$. On the other hand  sigmoid loss, 
ramp loss retain accuracy of at least $96\%$ even under $40\%$ noise. Also under non-uniform noise and class conditional 
noise, accuracies of sigmoid loss and  ramp loss are around $98\%$ whereas accuracy of SVM is only $77\%$. It is easy to see 
the noise tolerance of risk minimization with sigmoid loss or ramp loss when compared to the performance of SVM.

\begin{figure}
 \begin{center}
  \begin{tabular}{cc}
 \includegraphics[scale=.3]{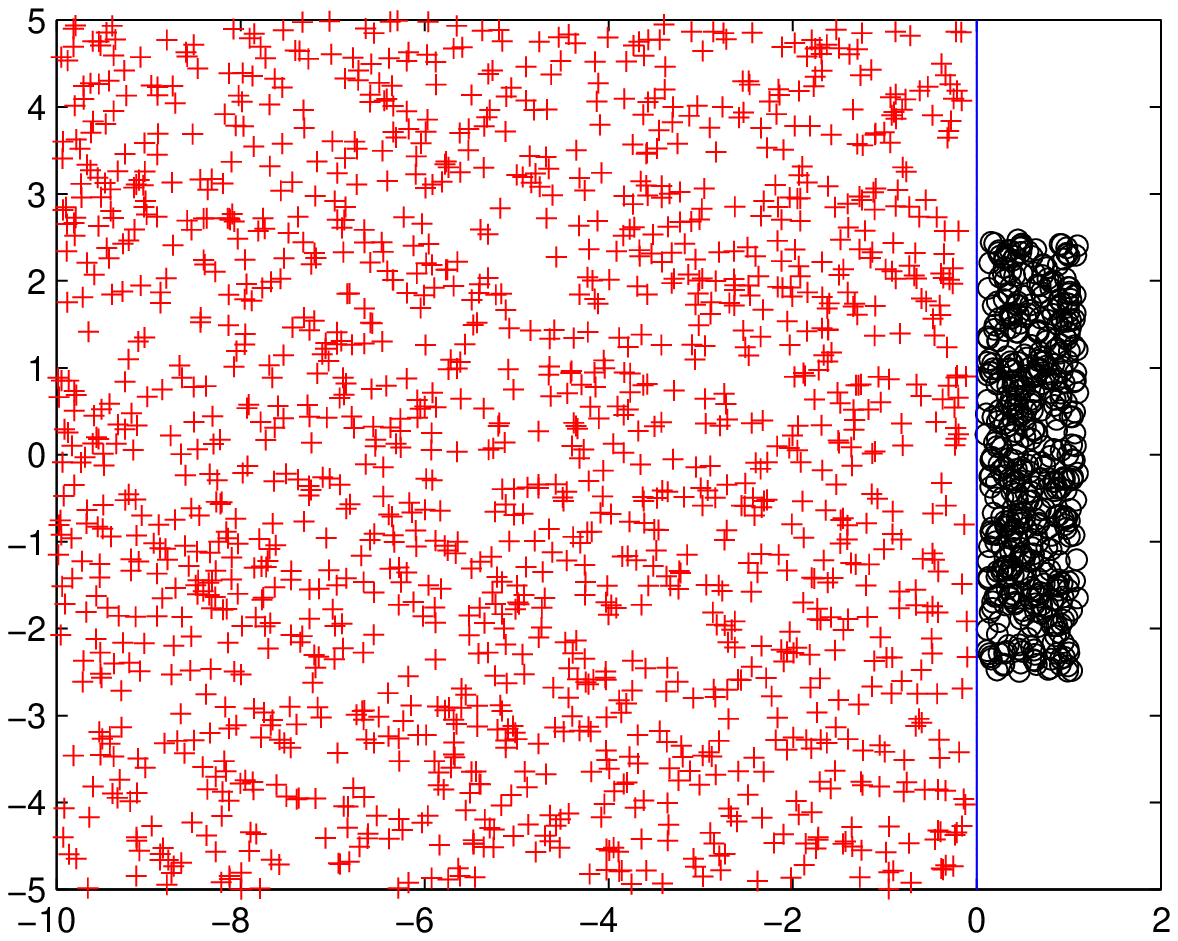}  &  \includegraphics[scale=.3]{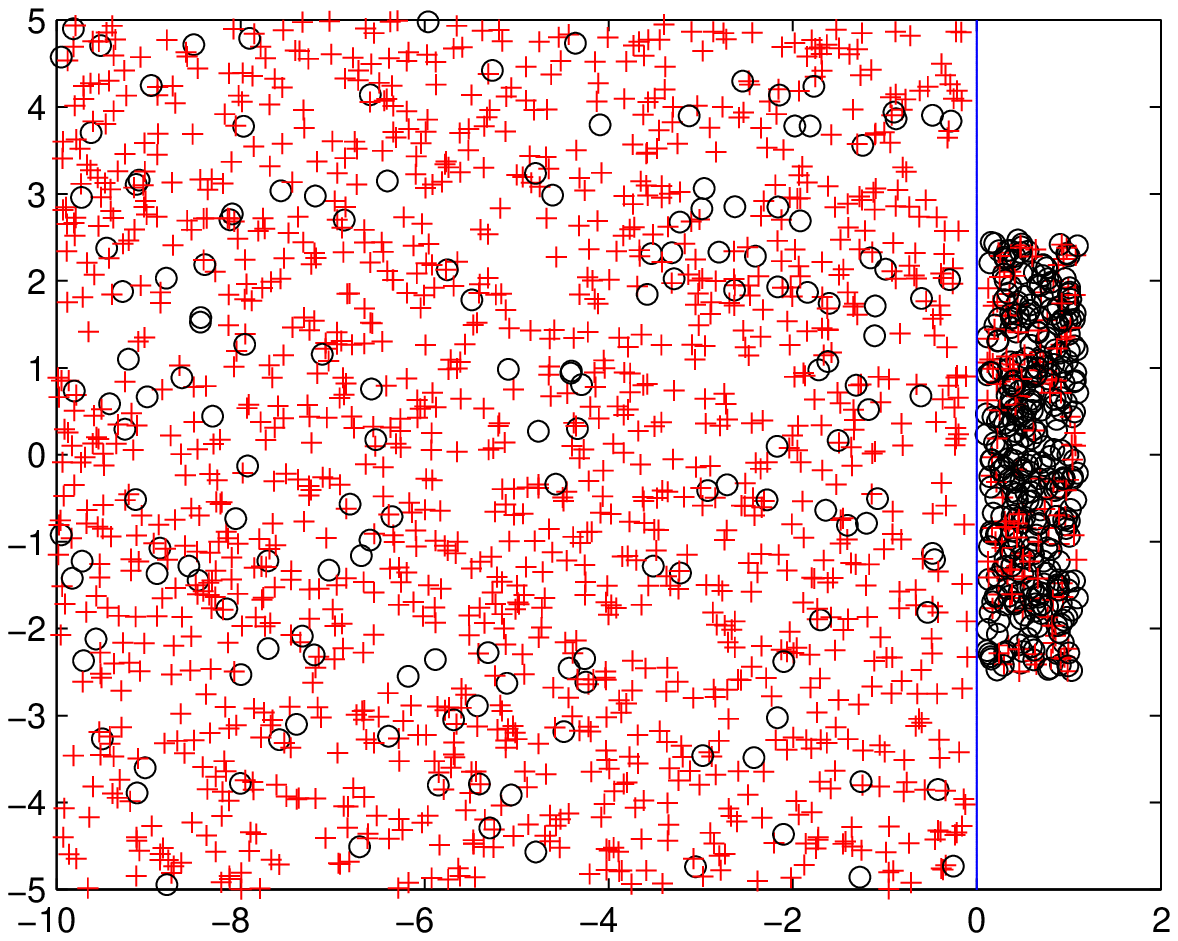}\\
(a)  &  (b) \\
\includegraphics[scale=.3]{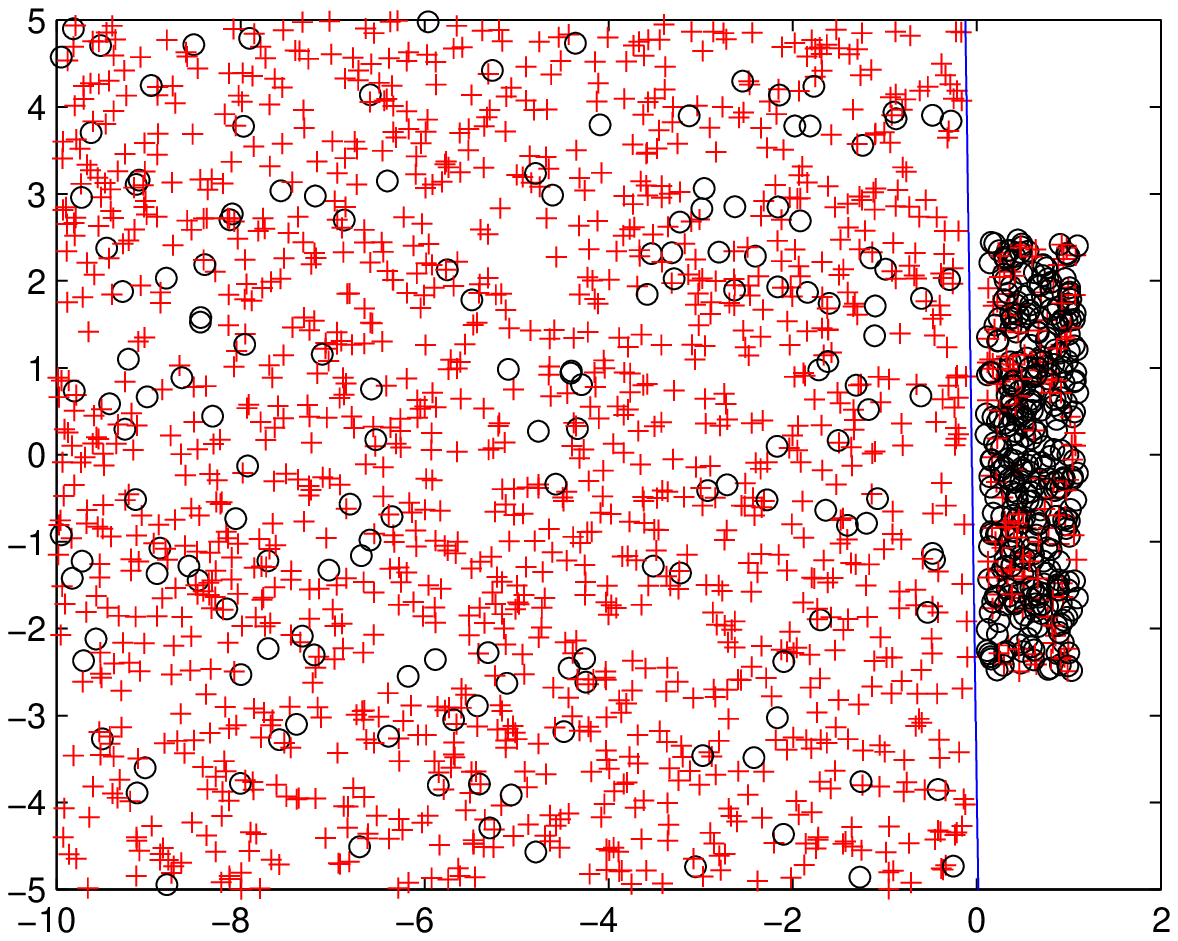}  &  \includegraphics[scale=.3]{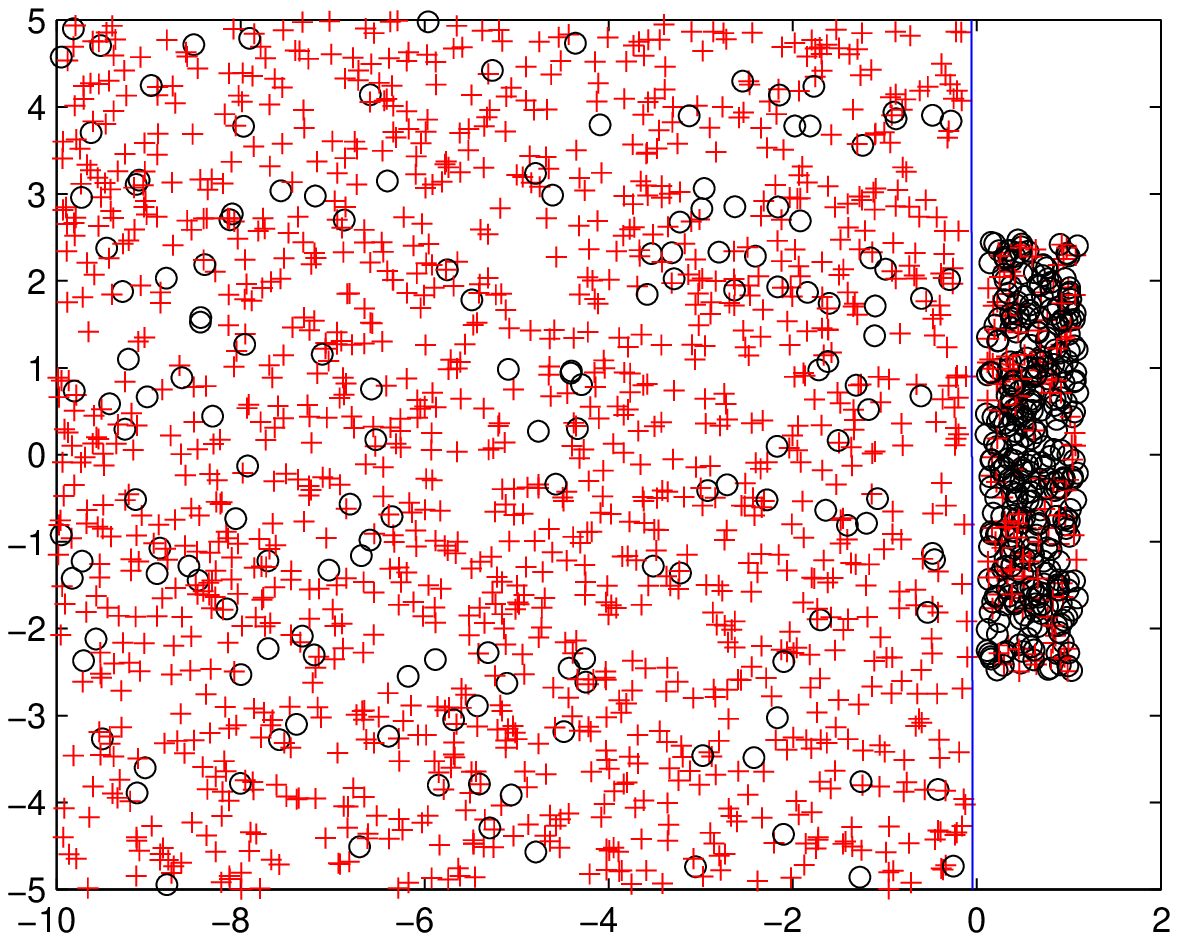}\\
(c)  &  (d) \\
\includegraphics[scale=0.3]{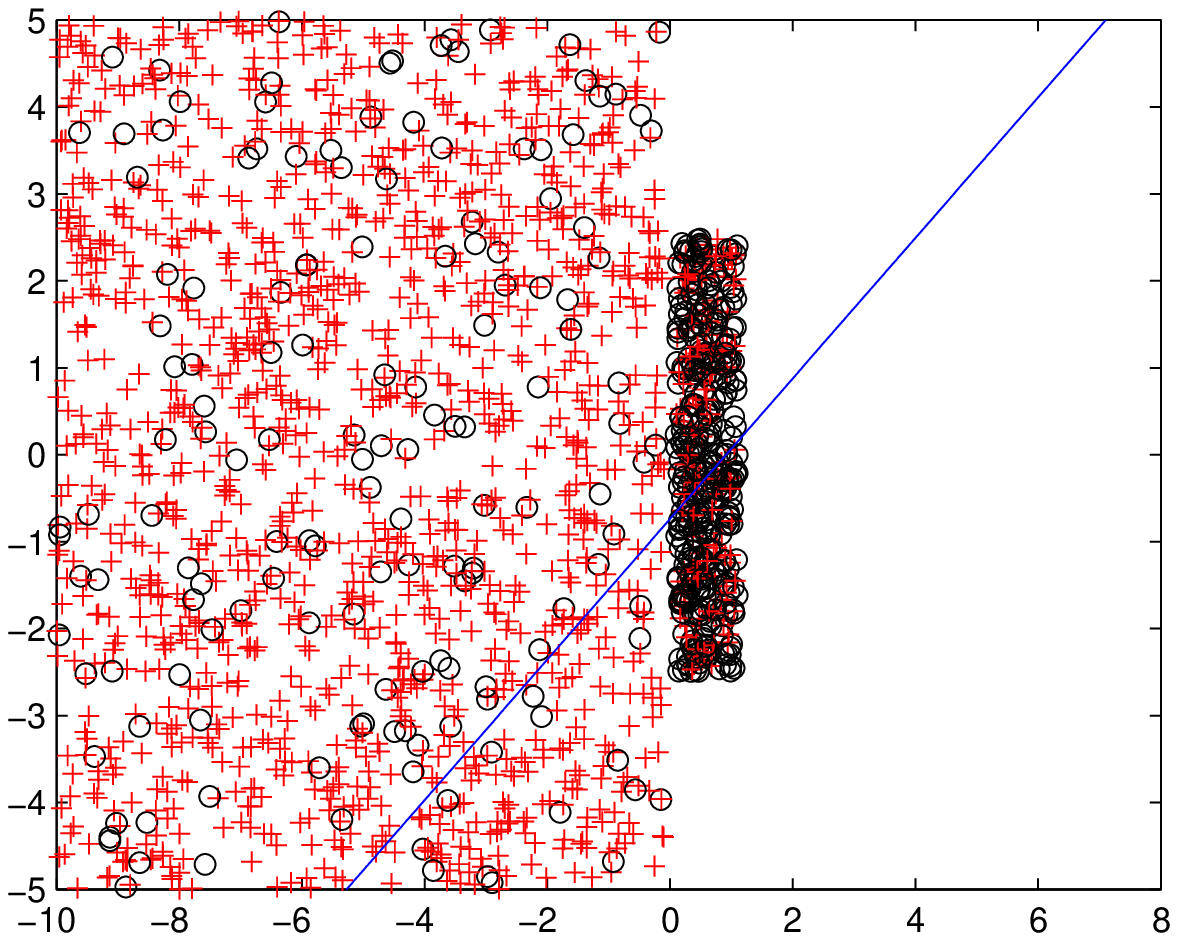} & \includegraphics[scale=0.3]{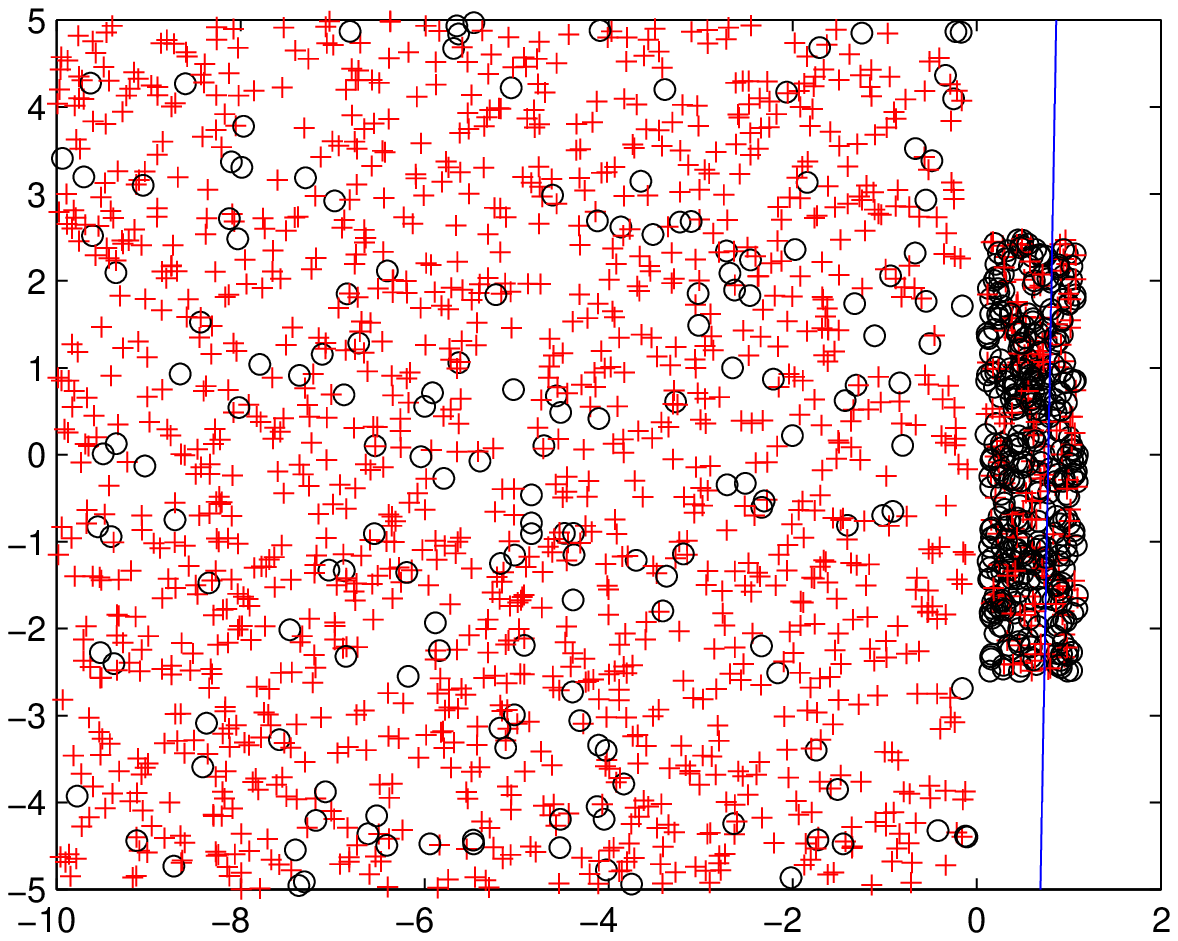}\\
(e) & (f)
  \end{tabular}
\caption{Results of different algorithms on Synthetic Dataset 3. (a) The data along with true classifier (Solid line),
(b) data corrupted with  class-conditional label noise with noise rates 30\% and 10\%. 
Linear classifiers learnt by minimizing (c) sigmoid loss (d)ramp loss
(e) hinge loss (linear SVM) (f) Square loss}
\label{fig:1d-example7}
 \end{center}
\end{figure}

\begin{figure}
\begin{center}
\includegraphics[scale=0.17]{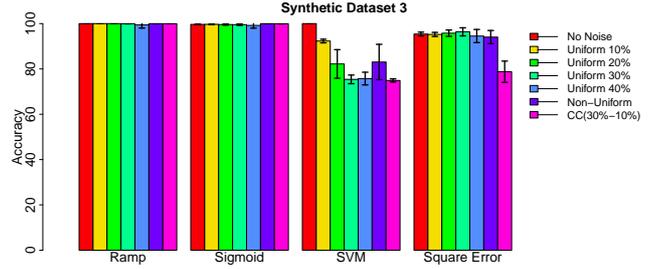}
\caption{Comparison Results on Synthetic Dataset 3}
\label{P3}
\end{center}
\end{figure}

In Synthetic Dataset 3, we have imbalanced set of training examples and asymmetric class regions in $\R^2$. 
But here, we have uniform class conditional densities. Figure~\ref{fig:1d-example7} shows classifiers learnt using sigmoid loss, ramp loss, hinge loss and square error loss on Synthetic Dataset~3 with class conditional label noise.
Here again, we see that sigmoid loss and ramp loss based approaches correctly find the true classifier.
Whereas, hinge loss and square error loss based approaches fail to learn the true classifier.
As can be seen from Figure~\ref{P3}, under $10\%$ uniform noise, accuracy of SVM drops to $92.3\%$. 
Then it decreases to $75.8\%$ under $40\%$ uniform noise.  
Accuracies of sigmoid loss, ramp loss stay above $99\%$ even under $40\%$ noise. 
Under non-uniform noise and class conditional noise, both sigmoid loss and ramp loss outperform SVM.

\begin{figure}
\begin{center}
\includegraphics[scale=0.17]{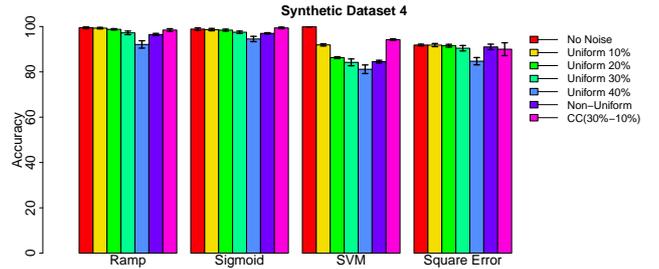}
\caption{Comparison Results on Synthetic Dataset 4}
\label{P4}
\end{center}
\end{figure}

In Synthetic Dataset~4, we have imbalanced, asymmetric classes in $\R^{50}$. 
As can be seen from Figure~\ref{P4}, the performance of noise-tolerant loss functions stays good 
even in these higher dimensions. The figure also show that the SVM method is not robust to label noise and its accuracies 
keep dropping when there is label noise.

\begin{figure}[tp]
 \begin{center}
  \begin{tabular}{cc}
 \includegraphics[scale=.3]{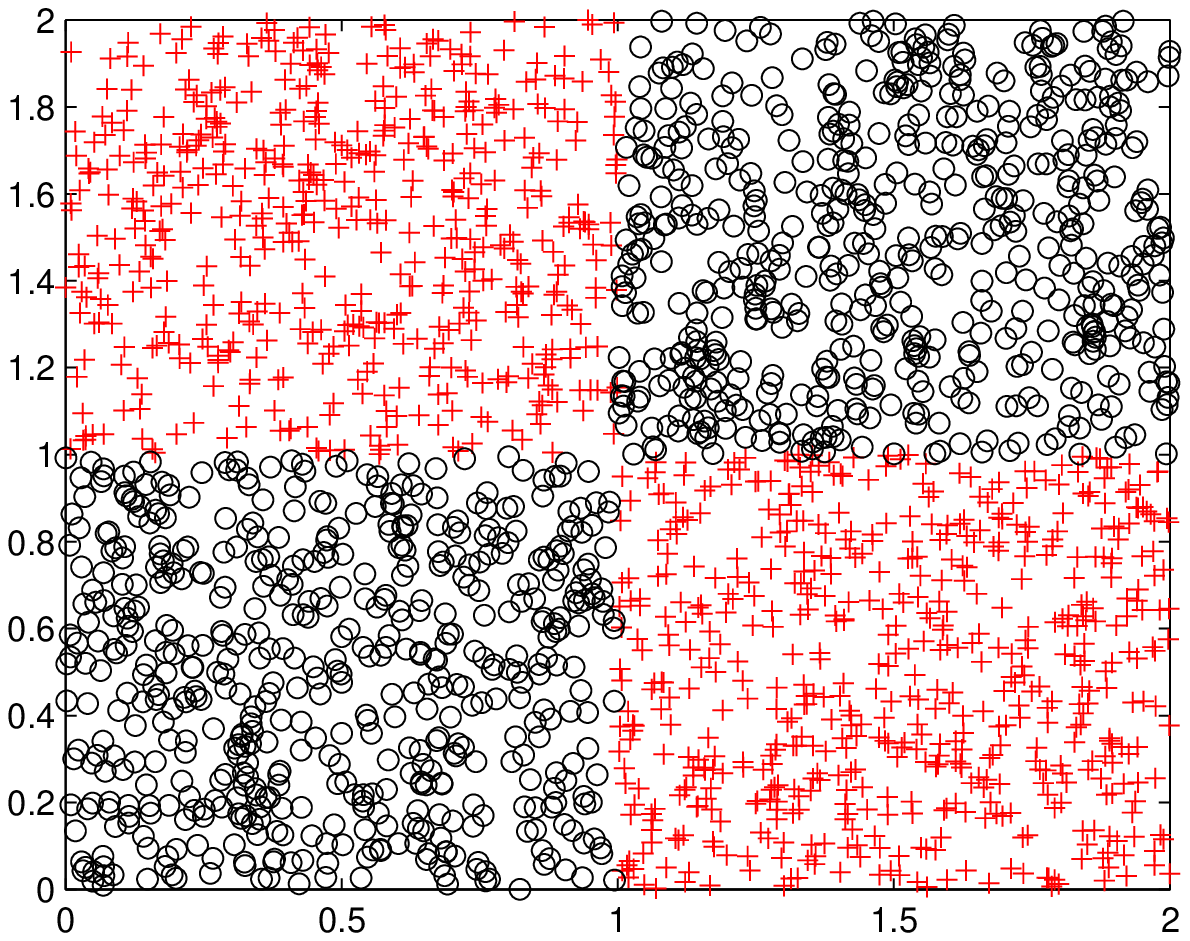}  &  \includegraphics[scale=.3]{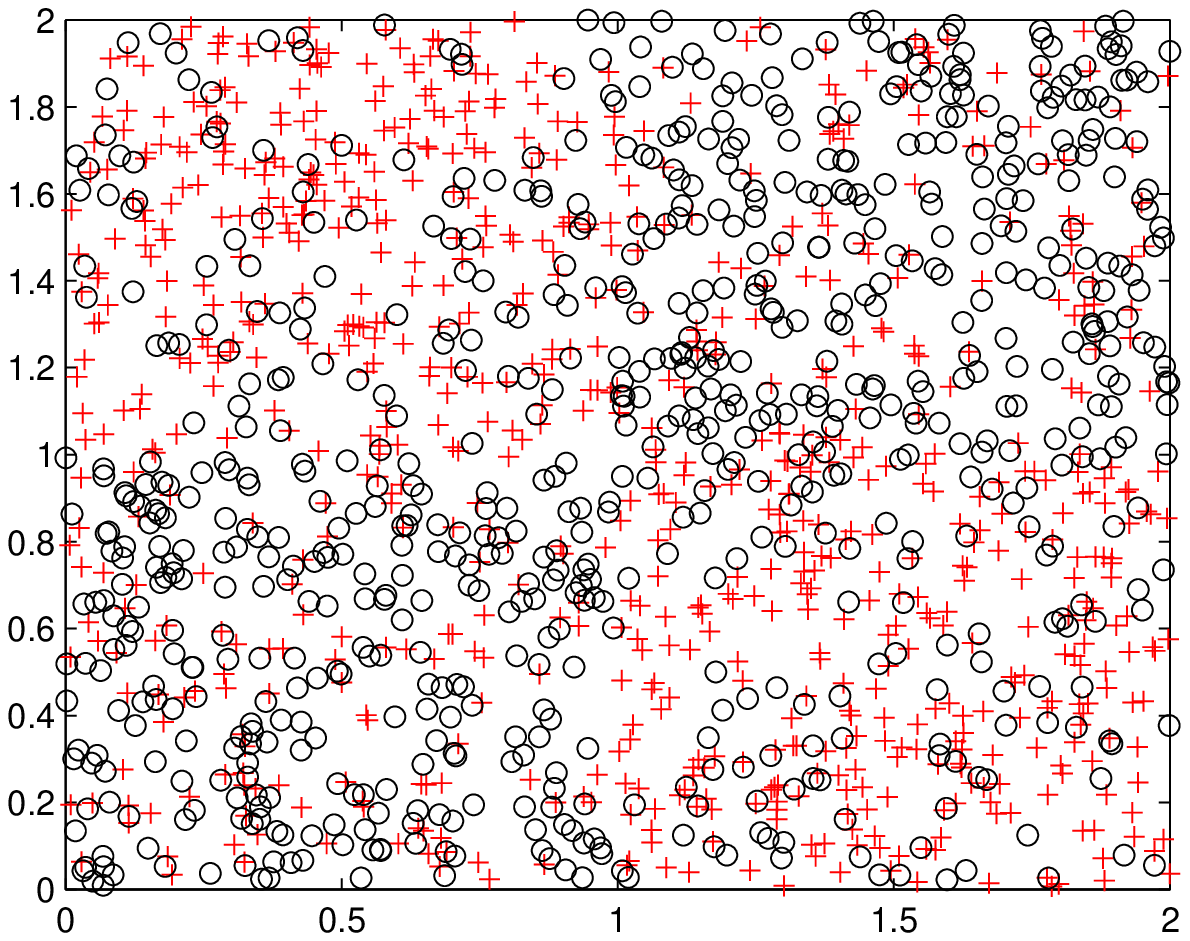}\\
(a)    &    (b)   \\
\includegraphics[scale=.3]{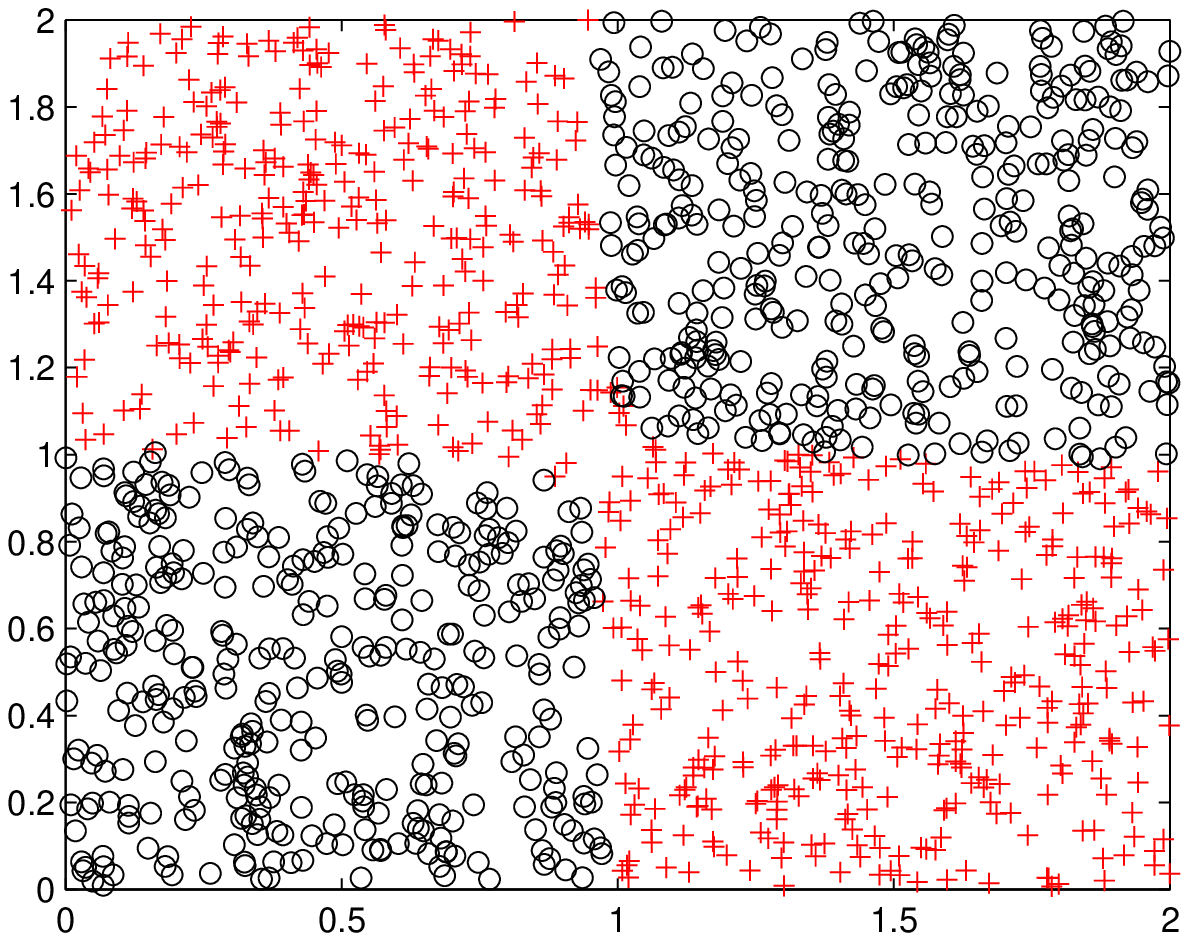}    &  \includegraphics[scale=.3]{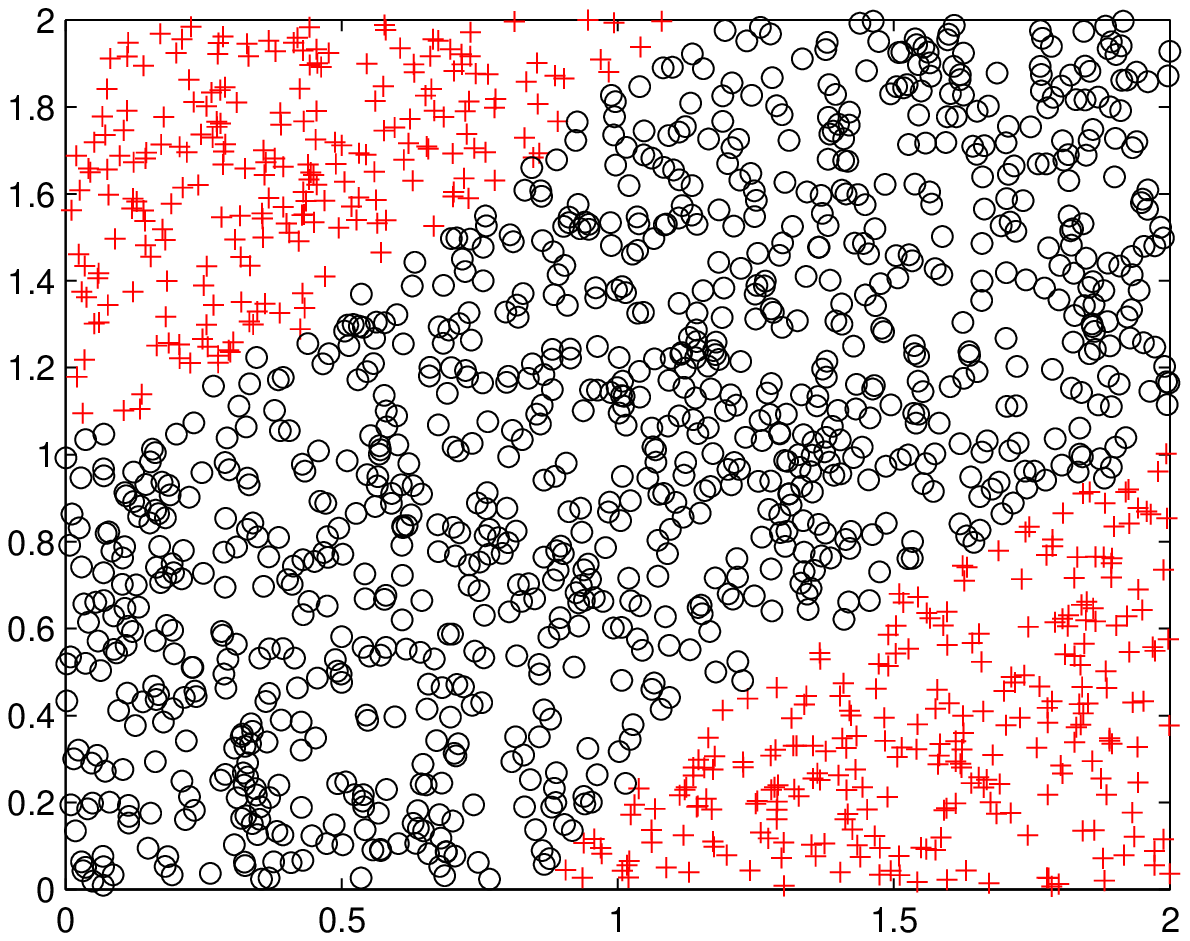}  \\
(c)  &  (d)  \\
  \end{tabular}
\caption{Results of different algorithms using quadratic kernel on Synthetic Dataset~5. (a) the data along with true class, 
(b) data corrupted with 30\% uniform noise. Labeling of quadratic classifiers learnt by minimizing (c) ramp loss (d) hinge loss.}
\label{fig:quad}
 \end{center}
\end{figure}

\begin{table}
\begin{center}
\caption{Comparison Results on Synthetic Dataset 5}
\label{P5}
\begin{tabular}{cccc}
\hline
\hline
Noise Rate & kernel & SVM & Ramp Loss\\ 
\hline
\hline
0\%    & quadratic  & 99.61$\pm$0.18  & 99.6$\pm$0.2 \\
Uni. 15\% & quadratic & 90.26$\pm$3.9 & 99.28$\pm$ 0.32 \\ 
Uni. 30\% & quadratic & 80.97$\pm$4.7 & 98.5$\pm$0.8\\
\hline
0\% & Gaussian & 98.93$\pm$0.6 & 98.9$\pm$0.6\\
Uni. 15\% & Gaussian & 96.3$\pm$0.6 & 99.06$\pm$0.9\\ 
Uni. 30\% & Gaussian & 93.6$\pm$1.7 & 96.3$\pm$1.1\\
\hline
\hline
\end{tabular}
\end{center}
\end{table}

Figure~\ref{fig:quad} shows the classifiers learnt using SVM and ramp loss on Synthetic Dataset 5 ($2 \times 2$ checker board) with 30\% label noise. Quadratic kernel is used in both approaches to capture the nonlinear classification boundary. We see that ramp loss based classifier accurately captures the true classifier, while
SVM completely misses it.
We can see in Table~\ref{P5}, on $2\times 2$-checker board data, accuracy of SVM with quadratic kernel drops to 90\% under 15\% noise and 80\% under 30\% noise from 99\% under noise free data. Ramp Loss shows impressive noise tolerance while using quadratic kernel. Ramp loss retains 98.5\% accuracy even under 30\% noise.
SVM with Gaussian kernel achieves better noise tolerance than SVM with polynomial kernel on 2$\times$2-checker board data. Accuracy of SVM drops to 93.6\%(80\%) under 30\% noise when using Gaussian(quadratic) kernel. Ramp loss performs better, retaining 96.3\% accuracy under 30\% noise. 

\subsubsection{Results on UCI Datasets}
We now discuss the performances on the 5 benchmark data sets from UCI ML repository. 
On the Ionosphere data the 
accuracy achieved by a linear classifier (even in noise-free case) is high. We compare risk minimization with sigmoid and ramp loss 
on this data against the performance of SVM under uniform noise.
On Ionosphere dataset, as can be seen from Table~\ref{Ion}, accuracy of SVM drops to $70.3\%$ under $40\%$ noise from $85\%$ 
under no-noise, whereas Ramp loss drops to $75.1\%$ from $84.7\%$. Sigmoid loss performs similar to Ramp loss.

\begin{table}
\begin{center}
\caption{Comparison Results on Ionosphere Dataset}
\label{Ion}
\begin{tabular}{ccccc}
\hline
\hline
Noise($\eta$) & Ramp & Sigmoid & SVM & Sq.Err. \\
\hline
\hline
0\% &$84.7\pm 2.8$&$83.1\pm 3.6 $&$85.2\pm 3.8$&$85.6\pm 2.8$\\
Uni 10\%&$83.1\pm 3.1$&$82.4\pm 3.3	$&$82.75\pm 4.2$&$84.9\pm 2.7$\\
Uni 20\%&$81.2\pm 3.9$&$81.8\pm 4.1 $&$79\pm 3.8$&$81.9\pm 4.4$\\
Uni 30\%&$77.7\pm 4.4$&$77.1\pm 5.1 $&$76.1\pm 5.5$&$77.7\pm 5.1$\\
Uni 40\%&$75.1\pm 4.1$&$74.2\pm 6.8 $&$70.3\pm 4.9$&$69.2\pm 5.95$\\
\hline
\hline
\end{tabular}
\end{center}
\end{table}

\begin{table}
\begin{center}
\caption{Results on Balance, Heart and Vote Datasets Using Gaussian Kernel}
\label{gauss}
\begin{tabular}{cccc}
\hline
\hline
Dataset & Noise Rate & SVM & Ramp Loss\\
\hline
\hline
\multirow{3}{*}{\parbox{1.5cm}{Balance}} &  0\% & 99.30$\pm$1.16 & 99.30$\pm$ 1.2 \\
& Uni 15\% & 96.06$\pm$2.4& 97.7$\pm$1.17\\ 
& Uni 30\% & 82.1$\pm$11.2& 92.1$\pm$7.4\\\hline
\multirow{3}{*}{\parbox{1.5cm}{Heart}} & 0\% & 82.58$\pm$7.82 & 83.33$\pm$4.56 \\
& Uni 15\% & 80.6$\pm$8.85& 84.07$\pm$ 7.10\\ 
& Uni 30\% & 77.36$\pm$ 9.31& 79.10$\pm$9.94\\ \hline
\multirow{3}{*}{\parbox{1.5cm}{Vote}} & 0\% &94.49$\pm$1.64 &94.49$\pm$1.64  \\
& Uni 15\% & 90.67$\pm$4.4& 90.36$\pm$4.2\\ 
& Uni 30\% & 81.2$\pm$5.8& 85.32$\pm$6.7\\
\hline
\hline
\end{tabular}
\end{center}
\end{table}

 On Balance, Heart and Vote datasets, we explore SVM and Ramp loss using Gaussian kernel under uniform noise. The results on these three datasets are described in Table~\ref{gauss}.
We can see in Balance dataset, in Table~\ref{gauss}, accuracy of SVM drops to 82\% under 30\% noise from 99\% on noise free data while ramp loss retains 92\% accuracy. In Heart dataset, ramp loss performs better than SVM. In Vote dataset, performance of Ramp loss is marginally better.

\begin{table*}
\begin{center}
\caption{Comparison Results on Breast Cancer Dataset}
\label{Bre}
\begin{tabular}{cccccc}
\hline
\hline
Noise($\eta$) & Ramp & Sigmoid & SVM & Sq.Err. & $0-1$\\
\hline
\hline
0\%  & $97.7\pm1.6$&$97.8\pm1.5$&$96.8\pm0.6$&$97.4\pm0.4$ &$95.8\pm 1.3$\\
Uniform 10\%  &$ 97.5\pm1.7$&$97.7\pm1.6$&$96.7\pm0.7$&$97.34\pm 1.8$&$96.4\pm 1$\\
Uniform 20\%  &  $97.1\pm 1.7$&$97.05\pm 1.7$&$96.3\pm0.9$&$96.9\pm1.7$&$96.3\pm0.9$\\
Uniform 30\%  & $96.1\pm 2.2$&$96.05\pm 2.9$&$94.3\pm 3.08$ & $94.26 \pm 3.6$&$96.2\pm 1.5$\\
Uniform 40\% & $93.2\pm 4.8$   &$ 92.6\pm 4.1$ & $  88.8\pm 4.7$ &$   88.1\pm 6.7$&$93.5\pm 2.8$\\
Non Uniform  & $ 94.4\pm 1.2 $  & $ 93.2\pm 1.7 $ & $92.8\pm 3.5 $  & $92.4\pm 2.3 $ &  $95.9\pm0.9  $  \\
CC (40\%-20\%) & $89.4\pm 2.4 $&$89.1\pm 3.2$& $86.1\pm 7.4$ & $86.24\pm  4.2$&$95.4\pm0.6$ \\ \hline
\hline
\end{tabular}
\end{center}
\end{table*}

Breast Cancer data set has almost separable classes 
and a linear classifier performs well. On Breast Cancer data set we 
compare $0-1$ loss, sigmoid loss and ramp loss with SVM (hinge loss).
In breast cancer problem, as can be seen in Table~\ref{Bre}, accuracy of CALA algorithm drops to $93.5\%$ under $40\%$ noise 
from $95.8\%$ under no-noise. Sigmoid loss and Ramp loss drop to $93\%$ under $40\%$ noise. 
Accuracy of SVM drops to $89\%$ under $40\%$ noise.  
Under non-uniform noise and class conditional noise, risk minimization under $0$-$1$ loss, Sigmoid loss, Ramp loss perform better than SVM.

All the results presented here, amply demonstrate the noise tolerance of risk minimization under sigmoid loss and ramp loss 
which satisfy our theoretical conditions for noise tolerance. In contrast, the SVM method does not exhibit much robustness 
to label noise. 
Using synthetic data sets we have demonstrated that SVM is particularly vulnerable to label 
noise under certain  kinds of geometry of pattern classes.  
Under balanced training set, symmetric classes with uniform densities, SVM performs moderately well under noise. 
But if we have intra-class nonuniform density or imbalanced training set  along with asymmetric class regions, 
then accuracy of SVM drops severely when training data are corrupted with label noise. This is demonstrated in two dimensions 
through problems 2 and 3 and in higher dimensions through problems 4. On the other hand 
risk minimization with $0$-$1$ loss, ramp loss and sigmoid loss exhibit impressive 
impressive noise tolerance abilities as can be seen from our results on synthetic as well as real data sets.

\section{Conclusions and Future Work}
\label{sec:conclusions}

In this paper, we  analyzed the noise tolerance of  risk minimization which is a generic method for learning classifiers. 
We derived some sufficient conditions on a loss function for risk minimization under that loss function to be noise tolerant under uniform 
and non-uniform label noise.
It is known $0-1$ loss is noise tolerant under uniform and non-uniform noise \cite{DBLP:journals/tcyb/ManwaniS13}. 
The result we presented here is generalization of that result. Our result shows that sigmoid loss, ramp loss and probit loss are all 
noise tolerant under uniform label noise. We also presented results to show that risk minimization under these loss 
functions can be noise tolerant to non-uniform label noise also if a parameter in the loss function is sufficiently high. 
Our theoretical results provide justification for the known superiority of the ramp loss over SVM in empirical studies.  
We also generalized a result on noise tolerance of $0-1$ loss under class conditional label noise proved in 
to the case of any loss function that satisfies a sufficient condition. This shows that sigmoid loss, ramp loss etc. can be 
used for noise robust learning of classifiers under class conditional noise. 

Through extensive empirical studies we demonstrated the noise tolerance of sigmoid loss, ramp loss and $0-1$ loss and also showed that 
the popular SVM method is not robust to label noise. We also showed specific types of class geometries in 2-class problem that 
make SVM sensitive to label noise.

All these noise tolerant losses are non-convex which makes the risk minimization harder. 
Risk minimization under $0-1$ loss is known to be hard. But the sigmoid loss, ramp loss etc are smooth and hence 
here we have used simple gradient descent for risk minimization under these loss functions. But, in general, such an approach would 
not be efficient to learn nonlinear classifiers under these losses. 
To do that, we have derived a DC program based risk minimization 
algorithm for ramp loss. For ramp loss, this approach allows to use kernel functions by default. Thus, making it easy to learn robust nonlinear classifiers.   

We can extend the concept of noise tolerance by introducing degree of noise tolerance.
Degree of noise tolerance could be defined as the difference of misclassification probability $f_{\eta}^*$
and $f^*$ on noise free data. 
$0-1$ loss, ramp loss and sigmoid loss have highest degree of noise tolerance as the above difference is zero. Hence an interesting direction of work is 
to analyze different convex loss functions from the point of view of degree of noise tolerance.

\appendix
\section{Regularized Empirical Risk Minimization under Ramp Loss using DC Program}
\label{appdx:1}
\label{algo:ramp}
Ramp loss can be written as difference of two convex function.
\[L_{\mbox{ramp}}(f(\xx),y_{\xx})=[1-y_{\xx}f(\xx)]_{+}-[-1-y_{\xx}f(\xx)]_{+}\]
For a nonlinear classifier parameterized by $\ww$ as $f(\xx)=(\ww^T\phi(\xx)+b)$, the regularized empirical risk under ramp loss is
\begin{flalign*}
& R_{ramp}^{reg}(\ww,b)= \frac{1}{2}||\ww||^2+C\sum\limits_{i=1}^{N} [1-y_{\xx_i}(\ww^T\phi(\xx_i)+b)]_{+}\\
& -C\sum\limits_{i=1}^{N} [-1-y_{\xx_i}(\ww^T\phi(\xx_i)+b)]_{+}
\end{flalign*}
where $C$ is the regularization parameter and $\phi$ is a nonlinear transformation.
Let $\Theta=(\ww\;,b)$. $R^{reg}_{ramp}(\Theta)$ can be written as difference of two convex functions $Q_1(\Theta)$ and $Q_2(\Theta)$ where
\begin{eqnarray}
\nonumber Q_1(\Theta) &=&  \frac{1}{2}||\ww||^2+C\sum\limits_{i=1}^{N} [1-y_{\xx_i}(\ww^T\phi(\xx_i)+b)]_{+}\\
\nonumber Q_2(\Theta) &=&  C\sum\limits_{i=1}^{N} [-1-y_{\xx_i}(\ww^T\phi(\xx_i)+b)]_{+}
\end{eqnarray}
This decomposition leads to an efficient algorithm for minimization of $R^{reg}_{ramp}(\Theta)$ 
using DC (difference of convex) program \citep{Pham1997,Wu07robusttruncated}.
Here, we present the derivation of the DC program for minimizing $R^{reg}_{ramp}(\Theta)$. This algorithm is slightly different from the one discussed in
\citet{Wu07robusttruncated}. 
The high level DC program for minimizing
$R^{reg}_{ramp}(\Theta)$ is presented in Algorithm~\ref{algo1}. 
\begin{algorithm}
\caption{DC Algorithm for Minimizing $R_{reg}(\Theta)$}
\label{algo1}
{\bf Initialize} $\Theta^{(0)}$\;
\Repeat{convergence of $\Theta^{(l)}$}{
\[\Theta^{(l+1)}=\arg\min_{\Theta}\;   Q_1(\Theta)- \Theta^T \nabla Q_2(\Theta^{(l)})\]
}
\end{algorithm}
We initialize $\Theta^{(0)}$ as $\Theta^{(0)} = \arg\min\limits_{\ww,b} Q_1(\Theta)$.
That is, we find the SVM classifier and initialize with that.
Now we derive the main step of the DC program for finding $\Theta^{(l+1)}$.
\subsection{Finding $\Theta^{(l+1)}$}
Given $\Theta^{(l)}$, $\Theta^{(l+1)}$ is found as
 \[\Theta^{(l+1)}=\arg\min_{\Theta}\;   Q_1(\Theta)- \Theta^T \nabla Q_2(\Theta^{(l)})\]
We note that
\[\nabla Q_2(\Theta^{(l)}) = \Big{[}-\sum\limits_{i=1}^N  \beta_i^{(l)} y_{\xx_i}\phi(\xx_i)^T \;\;\;
-\sum\limits_{i=1}^N  \beta_i^{(l)} y_{\xx_i}\Big{]}^T \]
where $\beta_i^{(l)}=\mathbb{I}_{\{y_{\xx_i}(\phi(\xx_i)^T\ww^{(l)}+b^{(l)})<-1\}} C$.
The second step of DC program has the following form
\begin{flalign*}
\nonumber &\Theta_{t+1} = \arg\min\limits_{\Theta}\;\; Q_1(\Theta)-\Theta^T \nabla Q_2(\Theta^{(l)})\\
\nonumber &= \arg\min\limits_{\ww,b,\xi} \;\; \frac{1}{2}||\ww||^2+C\sum\limits_{i=1}^N \xi_i +\sum\limits_{i=1}^N \beta_i^{(l)} y_{\xx_i} (\ww^T\phi(\xx_i)+b)\\
\nonumber & \mbox{s.t.} \;\; \xi_i \geq 0, \;\;(\ww^T\phi(\xx_i)+b)y_{\xx_i}\geq 1-\xi_i,\;\;
i=1\ldots N
\end{flalign*}
where $\beta_i^{(l)}=C$ if $y_{\xx_i}(\phi(\xx_i)^T\ww^{(l)}+b^{(l)})<-1$, and $\beta_i^{(l)}=0$ otherwise.
It is to be noted $\beta_i^{(l)}$ depends on $\Theta^{(l)}$. The Lagrangian will be
\begin{flalign*}
&L(\ww,b,\xi) = \frac{1}{2}||\ww||^2+C\sum\limits_{i=1}^N \xi_i +\sum\limits_{i=1}^N \beta_i^{(l)} y_{\xx_i} (\ww^T\phi(\xx_i)+b) \\
&\;\;\;\;-\sum\limits_{i=1}^N \mu_i \xi_i-\sum\limits_{i=1}^N \alpha_i[y_{\xx_i}(\ww^T\phi(\xx_i)+b)-1+\xi_i]
\end{flalign*}
Now the dual optimization problem is
\begin{eqnarray}
\max_{\alphaa,\muu} \min\limits_{\ww,b,\xi}  & L(\ww,b,\xi) \nonumber \\
  \mbox{s.t.} & \alpha_i \geq 0,\; \mu_i\geq 0,\; i=1\ldots N \nonumber 
\end{eqnarray}
where $\alphaa = [\alpha_1 \; \alpha_2\;\cdots \; \alpha_N]$ and $\muu = [\mu_1\;\mu_2\; \cdots \; \mu_N]$. The partial derivatives of $L$ with respect to the $\ww,\;b$ and $\xi_i$ are
\begin{eqnarray}
\nonumber \frac{\partial L}{\partial \ww} &=& \ww-\sum\limits_{i=1}^N y_{\xx_i}\phi(\xx_i)(\alpha_i-\beta_i^{(l)})=0 \\
\nonumber \frac{\partial L}{\partial b}   &=& \sum\limits_{i=1}^N (\alpha_i-\beta_i^{(l)})y_{\xx_i}=0\\
\nonumber \frac{\partial L}{\partial \xi_i} &=& C-\mu_i-\alpha_i=0,\;\; i=1\ldots N
\end{eqnarray}
Complementary slackness conditions are,
\[\mu_i\xi_i=0,\;\;\alpha_i[y_{\xx_i}(\ww^T\phi(\xx_i)+b)-1+\xi_i]=0,\;\;\; i=1\ldots N\]
The Wolf dual will become
\begin{eqnarray}
\nonumber \max\limits_{\alphaa} & \sum\limits_{i=1}^N\alpha_i-\frac{1}{2}||\sum\limits_{i=1}^N (\alpha_i-\beta_i^{(l)})\phi(\xx_i)y_{\xx_i}||^2_2 \\
\nonumber \mbox{s.t.} &\sum\limits_{i=1}^N(\alpha_i-\beta_i^{(l)})y_{\xx_i}=0 \\
\nonumber & 0 \leq \alpha_i \leq C,\;\; i=1\ldots N
\end{eqnarray}
We can simplify it further. Let $\lambda_i=(\alpha_i-\beta_i^{(l)}),\; i=1\ldots N$. Now the dual will become,
\begin{eqnarray}
\nonumber  \max\limits_{\lambdaa} & \sum\limits_{i=1}^N \lambda_i-\frac{1}{2}||\sum\limits_{i=1}^N \lambda_i\phi(\xx_i)y_{\xx_i}||^2_2+\mbox{const}\\
\nonumber \mbox{s.t.} & \sum\limits_{i=1}^N \lambda_i y_{\xx_i}=0 \\
\nonumber & 0 \leq \lambda_i \leq C,\;\; \forall i \mbox{ s.t. } \beta_i^{(l)}=0\\
\nonumber & -C\leq \lambda_i \leq 0,\;\; \forall i \mbox{ s.t. } \beta_i^{(l)}=C
\end{eqnarray}
where $\lambdaa=[\lambda_1\; \lambda_2 \; \cdots \; \lambda_N]$. 
Let $V^{(l+1)} = \{\; i \; | \; -\beta_i^{(l)}<\lambda_i^{(l+1)}<C-\beta_i^{(l)} \;\}$.
Find $\Theta^{(l+1)} = (\ww^{(l+1)},b^{(l+1)})$ using,
\begin{eqnarray}
\nonumber \ww^{(l+1)} &=& \sum\limits_{i=1}^N \lambda_i^{(l+1)} \phi(\xx_i) y_{\xx_i}\\
\nonumber  b^{(l+1)}  &=&  \frac{1}{|V^{(l+1)}|}
\sum_{i \in V^{(l+1)}}[y_{\xx_i}-\phi(\xx_i)^T\ww^{(l+1)}]
\end{eqnarray}
For minimizing the quadratic program, we use generalized sequential minimal optimization \citep{Keerthi:2002} for fast convergence.
The complete DC algorithm for learning a classifier
by minimizing $R^{reg}_{ramp}(\Theta)$ is described in Algorithm~\ref{algo2}. 

\begin{algorithm}[t]
\caption{DC Algorithm for Minimizing $R^{reg}_{ramp}(\Theta)$}
\label{algo2}
\KwIn{$C>0$, Training Dataset $\mathcal{S}$}
\KwOut{$\ww^*,b^*$}
\Begin
{
{\bf Initialize} $l=0$, $\Theta^{(0)} = \arg\min\limits_{\ww,b} \frac{1}{2}||\ww||^2+C\sum\limits_{i=1}^{N} [1-y_{\xx_i}(\ww^T\phi(\xx_i)+b)]_{+}$\;
\Repeat{convergence of $\Theta^{(l)}$}{
\begin{enumerate}
\item Find $\beta_i^{(l)},\;i=1\ldots N$ as
\[\beta_i^{(l)}=\mathbb{I}_{\{y_{\xx_i}(\phi(\xx_i)^T\ww^{(l)}+b^{(l)})<-1\}} C\]
\item Solve for $\lambdaa^{(l+1)}$ as
\begin{eqnarray}
\nonumber  \max\limits_{\lambdaa} & \sum\limits_{i=1}^N \lambda_i-\frac{1}{2}||\sum\limits_{i=1}^N \lambda\phi(\xx_i)y_{\xx_i}||^2_2\\
\nonumber \mbox{s.t.} & \sum\limits_{i=1}^N \lambda_i y_{\xx_i}=0 \\
\nonumber & 0 \leq \lambda_i \leq C,\;\; \forall i \mbox{ s.t. } \beta_i^{(l)}=0\\
\nonumber & -C\leq \lambda_i \leq 0,\;\; \forall i \mbox{ s.t. } \beta_i^{(l)}=C
\end{eqnarray}
\item Find $V^{(l+1)} = \{\; i\;|\;-\beta_i^{(l)}<\lambda_i^{(l+1)}<C-\beta_i^{(l)}\;\}$. Find $\Theta^{(l+1)} = (\ww^{(l+1)},b^{(l+1)})$ using,
\begin{eqnarray}
\nonumber \ww^{(l+1)} &=& \sum\limits_{i=1}^N \lambda_i^{(l+1)} \phi(\xx_i) y_{\xx_i}\\
\nonumber  b^{(l+1)}  &=&  \frac{1}{|V^{(l+1)}|}
\sum_{i \in V^{(l+1)}}[y_{\xx_i}-\phi(\xx_i)^T\ww^{(l+1)}]
\end{eqnarray}
\end{enumerate}
}
}
\end{algorithm}

\bibliography{me1}

\begin{thebibliography}{40}
\expandafter\ifx\csname natexlab\endcsname\relax\def\natexlab#1{#1}\fi
\providecommand{\bibinfo}[2]{#2}
\ifx\xfnm\relax \def\xfnm[#1]{\unskip,\space#1}\fi
\bibitem[{Fr\'{e}nay and Verleysen(2014)}]{frenay2013classification}
\bibinfo{author}{B.~Fr\'{e}nay}, \bibinfo{author}{M.~Verleysen},
\newblock \bibinfo{title}{{Classification in the Presence of Label Noise: A
  Survey}},
\newblock \bibinfo{journal}{IEEE Transactions on Neural Networks and Learning
  Systems} \bibinfo{volume}{25} (\bibinfo{year}{2014})
  \bibinfo{pages}{845--869}.
\bibitem[{Haussler(1992)}]{Hau1992}
\bibinfo{author}{D.~Haussler},
\newblock \bibinfo{title}{Decision theoretic generalizations of the {PAC} model
  for neural net and other learning applications},
\newblock \bibinfo{journal}{Information and Computation, Elsevier}
  \bibinfo{volume}{100} (\bibinfo{year}{1992}) \bibinfo{pages}{78--150}.
\bibitem[{Devroye et~al.(1996)Devroye, Gyorfi, and Lugosi}]{DGL1996}
\bibinfo{author}{L.~Devroye}, \bibinfo{author}{L.~Gyorfi},
  \bibinfo{author}{G.~Lugosi}, \bibinfo{title}{A Probabilistic Theory of
  Pattern Recognition}, \bibinfo{publisher}{Springer-Verlag},
  \bibinfo{address}{New York}, \bibinfo{year}{1996}.
\bibitem[{Bartlett et~al.(2006)Bartlett, Jordan, and McAuliffe}]{bjm-ccrb-05}
\bibinfo{author}{P.~L. Bartlett}, \bibinfo{author}{M.~I. Jordan},
  \bibinfo{author}{J.~D. McAuliffe},
\newblock \bibinfo{title}{Convexity, classification and risk bounds},
\newblock \bibinfo{journal}{Journal of the American Statistical Association}
  \bibinfo{volume}{101} (\bibinfo{year}{2006}) \bibinfo{pages}{138--156}.
\bibitem[{Manwani and Sastry(2013)}]{DBLP:journals/tcyb/ManwaniS13}
\bibinfo{author}{N.~Manwani}, \bibinfo{author}{P.~S. Sastry},
\newblock \bibinfo{title}{Noise tolerance under risk minimization},
\newblock \bibinfo{journal}{IEEE Transactions on Cybernetics}
  \bibinfo{volume}{43} (\bibinfo{year}{2013}) \bibinfo{pages}{1146--1151}.
\bibitem[{Nettleton et~al.(2010)Nettleton, Orriols-Puig, and
  Fornells}]{NOF2010}
\bibinfo{author}{D.~F. Nettleton}, \bibinfo{author}{A.~Orriols-Puig},
  \bibinfo{author}{A.~Fornells},
\newblock \bibinfo{title}{A study of the effect of different types of noise on
  the precision of supervised learning techniques},
\newblock \bibinfo{journal}{Artificial Intelligence Review}
  \bibinfo{volume}{33} (\bibinfo{year}{2010}) \bibinfo{pages}{275--306}.
\bibitem[{Fine et~al.(1999)Fine, Gilad-bachrach, Mendelson, and
  Tishby}]{fine1999noise}
\bibinfo{author}{S.~Fine}, \bibinfo{author}{R.~Gilad-bachrach},
  \bibinfo{author}{S.~Mendelson}, \bibinfo{author}{N.~Tishby},
\newblock \bibinfo{title}{Noise tolerant learnability via the dual learning
  problem},
\newblock in: \bibinfo{booktitle}{Proceedings of NSCT, June 1999}.
\bibitem[{Angelova et~al.(2005)Angelova, Abu-Mostafa, and
  Perona}]{Angelova:2005:PTS:1068507.1068955}
\bibinfo{author}{A.~Angelova}, \bibinfo{author}{Y.~Abu-Mostafa},
  \bibinfo{author}{P.~Perona},
\newblock \bibinfo{title}{Pruning training sets for learning of object
  categories},
\newblock in: \bibinfo{booktitle}{Proceedings of IEEE Computer Society
  Conference on Computer Vision and Pattern Recognition (CVPR) 2005},
  \bibinfo{address}{Washington, DC, USA}, pp. \bibinfo{pages}{494--501}.
\bibitem[{Brodley and Friedl(1999)}]{Brodley99identifyingmislabeled}
\bibinfo{author}{C.~E. Brodley}, \bibinfo{author}{M.~A. Friedl},
\newblock \bibinfo{title}{Identifying mislabeled training data},
\newblock \bibinfo{journal}{Journal Of Artificial Intelligence Research}
  \bibinfo{volume}{11} (\bibinfo{year}{1999}) \bibinfo{pages}{131--167}.
\bibitem[{Zhu et~al.(2003)Zhu, Wu, and Chen}]{Zhu:2003}
\bibinfo{author}{X.~Zhu}, \bibinfo{author}{X.~Wu}, \bibinfo{author}{Q.~Chen},
\newblock \bibinfo{title}{Eliminating class noise in large datasets},
\newblock in: \bibinfo{booktitle}{Proceedings of the Twentieth International
  Conference on Machine Learning (ICML), August 2003},
  \bibinfo{address}{Washington, DC, USA}, pp. \bibinfo{pages}{920--927}.
\bibitem[{John(1995)}]{John95robustdecision}
\bibinfo{author}{G.~H. John},
\newblock \bibinfo{title}{Robust decision trees: Removing outliers from
  databases},
\newblock in: \bibinfo{booktitle}{Proceedings of Ist International Conference
  Knowledge Discovery and Data Mining (KDD), August 1995},
  \bibinfo{address}{Montreal, Quebec, Canada}, pp. \bibinfo{pages}{174--179}.
\bibitem[{Daza and Acuna(2007)}]{daza2007algorithm}
\bibinfo{author}{L.~Daza}, \bibinfo{author}{E.~Acuna},
\newblock \bibinfo{title}{An algorithm for detecting noise on supervised
  classification},
\newblock in: \bibinfo{booktitle}{Proceedings of the 1st World Conference on
  Engineering and Computer Science (WCECS), October 2007},
  \bibinfo{address}{San Francisco, USA}, pp. \bibinfo{pages}{701--706}.
\bibitem[{Karmaker and Kwek(2006)}]{Karmaker:2006}
\bibinfo{author}{A.~Karmaker}, \bibinfo{author}{S.~Kwek},
\newblock \bibinfo{title}{A boosting approach to remove class label noise},
\newblock \bibinfo{journal}{International Journal of Hybrid Intelligent
  Systems} \bibinfo{volume}{3} (\bibinfo{year}{2006})
  \bibinfo{pages}{169--177}.
\bibitem[{Har-Peled et~al.(2007)Har-Peled, Roth, and Zimak}]{Sariel2007}
\bibinfo{author}{S.~Har-Peled}, \bibinfo{author}{D.~Roth},
  \bibinfo{author}{D.~Zimak},
\newblock \bibinfo{title}{Maximum margin coresets for active and noise tolerant
  learning},
\newblock in: \bibinfo{booktitle}{Proceedings of the 20th International Joint
  Conference on Artificial Intelligence (IJCAI), January 2007},
  \bibinfo{address}{Hyderabad, India}, pp. \bibinfo{pages}{836--841}.
\bibitem[{Bouveyron and Girard(2009)}]{journals/pr/BouveyronG09}
\bibinfo{author}{C.~Bouveyron}, \bibinfo{author}{S.~Girard},
\newblock \bibinfo{title}{Robust supervised classification with mixture models:
  Learning from data with uncertain labels},
\newblock \bibinfo{journal}{Pattern Recognition} \bibinfo{volume}{42}
  (\bibinfo{year}{2009}) \bibinfo{pages}{2649--2658}.
\bibitem[{Khardon and Wachman(2007)}]{Khardon:2007:NTV:1248659.1248667}
\bibinfo{author}{R.~Khardon}, \bibinfo{author}{G.~Wachman},
\newblock \bibinfo{title}{Noise tolerant variants of the perceptron algorithm},
\newblock \bibinfo{journal}{Journal Of Machine Learning Research}
  \bibinfo{volume}{8} (\bibinfo{year}{2007}) \bibinfo{pages}{227--248}.
\bibitem[{R{\"a}tsch et~al.(1999)R{\"a}tsch, Onoda, and M{\"u}ller}]{ROM1999}
\bibinfo{author}{G.~R{\"a}tsch}, \bibinfo{author}{T.~Onoda},
  \bibinfo{author}{K.~R. M{\"u}ller},
\newblock \bibinfo{title}{Regularizing adaboost},
\newblock in: \bibinfo{booktitle}{Proceedings of Advances in Neural Information
  Processing Systems (NIPS), November 1999}, \bibinfo{address}{Denver, CO,
  USA}, pp. \bibinfo{pages}{564--570}.
\bibitem[{Jin et~al.(2003)Jin, Liu, Si, Carbonell, and Hauptmann}]{jin2003new}
\bibinfo{author}{R.~Jin}, \bibinfo{author}{Y.~Liu}, \bibinfo{author}{L.~Si},
  \bibinfo{author}{J.~G. Carbonell}, \bibinfo{author}{A.~Hauptmann},
\newblock \bibinfo{title}{A new boosting algorithm using input-dependent
  regularizer},
\newblock in: \bibinfo{booktitle}{Proceedings of Twentieth International
  Conference on Machine Learning (ICML), August 2003},
  \bibinfo{address}{Washington D.C.}
\bibitem[{Biggio et~al.(2011)Biggio, Nelson, and
  Laskov}]{DBLP:journals/jmlr/BiggioNL11}
\bibinfo{author}{B.~Biggio}, \bibinfo{author}{B.~Nelson},
  \bibinfo{author}{P.~Laskov},
\newblock \bibinfo{title}{Support vector machines under adversarial label
  noise},
\newblock in: \bibinfo{booktitle}{Proceedings of the Third Asian Conference on
  Machine Learning (ACML), November 2011}, \bibinfo{address}{Taoyuan, Taiwan},
  pp. \bibinfo{pages}{97--112}.
\bibitem[{Kearns(1998)}]{Kearns:98}
\bibinfo{author}{M.~Kearns},
\newblock \bibinfo{title}{Efficient noise-tolerant learning from statistical
  queries},
\newblock \bibinfo{journal}{Journal of the ACM} \bibinfo{volume}{45}
  (\bibinfo{year}{1998}) \bibinfo{pages}{983--1006}.
\bibitem[{H\"{o}ffgen and Simon(1992)}]{Hoffgen:1992}
\bibinfo{author}{K.-U. H\"{o}ffgen}, \bibinfo{author}{H.~U. Simon},
\newblock \bibinfo{title}{Robust trainability of single neurons},
\newblock in: \bibinfo{booktitle}{Proceedings of the 5th Annual Workshop on
  Computational Learning Theory (COLT), 1992}, \bibinfo{address}{Pittsburgh,
  Pennsylvania, USA}, pp. \bibinfo{pages}{428--439}.
\bibitem[{Bylander(1994)}]{Bylander94}
\bibinfo{author}{T.~Bylander},
\newblock \bibinfo{title}{Learning linear threshold functions in the presence
  of classification noise},
\newblock in: \bibinfo{booktitle}{In Proceedings of the 7th Annual Workshop on
  Computational Learning Theory (COLT), 1994}, \bibinfo{address}{New Brunswick,
  New Jersey, USA}, pp. \bibinfo{pages}{340--347}.
\bibitem[{Blum and Frieze(1996)}]{Blum:1996}
\bibinfo{author}{A.~Blum}, \bibinfo{author}{A.~Frieze},
\newblock \bibinfo{title}{A polynomial-time algorithm for learning noisy linear
  threshold functions},
\newblock in: \bibinfo{booktitle}{Proceedings of the 37th Annual Symposium on
  Foundations of Computer Science (FOCS), October 1996},
  \bibinfo{address}{Burlington, Vermont, USA}, pp. \bibinfo{pages}{330--338}.
\bibitem[{Cohen(1997)}]{cohen1997learning}
\bibinfo{author}{E.~Cohen},
\newblock \bibinfo{title}{Learning noisy perceptrons by a perceptron in
  polynomial time},
\newblock in: \bibinfo{booktitle}{Proceedings of 38th Annual Symposium on
  Foundations of Computer Science (FOCS), October 1997},
  \bibinfo{address}{Miami Beach, Florida, USA}, pp. \bibinfo{pages}{514--523}.
\bibitem[{Stempfel et~al.(2007)Stempfel, Ralaivola, and
  Denis}]{stempfel2007learning}
\bibinfo{author}{G.~Stempfel}, \bibinfo{author}{L.~Ralaivola},
  \bibinfo{author}{F.~Denis}, \bibinfo{title}{Learning from Noisy Data using
  Hyperplane Sampling and Sample Averages}, \bibinfo{type}{Technical Report}
  \bibinfo{number}{3564}, HAL-CNRS, \bibinfo{address}{France},
  \bibinfo{year}{2007}.
\bibitem[{Scott et~al.(2013)Scott, Blanchard, and Handy}]{conf/colt/ScottBH13}
\bibinfo{author}{C.~Scott}, \bibinfo{author}{G.~Blanchard},
  \bibinfo{author}{G.~Handy},
\newblock \bibinfo{title}{Classification with asymmetric label noise:
  Consistency and maximal denoising.},
\newblock in: \bibinfo{booktitle}{Conference On Learning Theory},
  volume~\bibinfo{volume}{30} of \textit{\bibinfo{series}{W\&CP}},
  \bibinfo{publisher}{JMLR}, \bibinfo{year}{2013}, pp.
  \bibinfo{pages}{489--511}.
\bibitem[{Natarajan et~al.(????)Natarajan, Dhillon, Ravikumar, and
  Tewari}]{NIPS2013_5073}
\bibinfo{author}{N.~Natarajan}, \bibinfo{author}{I.~Dhillon},
  \bibinfo{author}{P.~Ravikumar}, \bibinfo{author}{A.~Tewari},
\newblock \bibinfo{title}{Learning with noisy labels},
\newblock in: \bibinfo{booktitle}{Proceedings of Advances in Neural Information
  Processing Systems (NIPS), December 2013}, \bibinfo{address}{Nevada, United
  States}, pp. \bibinfo{pages}{1196--1204}.
\bibitem[{Stempfel and Ralaivola(2009)}]{Stempfel:2009}
\bibinfo{author}{G.~Stempfel}, \bibinfo{author}{L.~Ralaivola},
\newblock \bibinfo{title}{{Learning SVMs from Sloppily Labeled Data}},
\newblock in: \bibinfo{booktitle}{Proceedings of the 19th International
  Conference on Artificial Neural Networks (ICANN), September 2009},
  \bibinfo{address}{Limassol, Cyprus}, pp. \bibinfo{pages}{884--893}.
\bibitem[{Sastry et~al.(2010)Sastry, Nagendra, and
  Manwani}]{DBLP:journals/tsmc/SastryNM10}
\bibinfo{author}{P.~S. Sastry}, \bibinfo{author}{G.~D. Nagendra},
  \bibinfo{author}{N.~Manwani},
\newblock \bibinfo{title}{A team of continuous-action learning automata for
  noise-tolerant learning of half-spaces},
\newblock \bibinfo{journal}{IEEE Transactions on Systems, Man and Cybernetics,
  Part--B} \bibinfo{volume}{40} (\bibinfo{year}{2010}) \bibinfo{pages}{19--28}.
\bibitem[{Thathachar and Sastry(2003)}]{Thathachar:2003:NLA:1197439}
\bibinfo{author}{M.~A.~L. Thathachar}, \bibinfo{author}{P.~S. Sastry},
  \bibinfo{title}{Networks of Learning Automata: Techniques for Online
  Stochastic Optimization}, \bibinfo{publisher}{Springer-Verlag New York,
  Inc.}, \bibinfo{address}{Secaucus, NJ, USA}, \bibinfo{year}{2003}.
\bibitem[{Brooks(2011)}]{DBLP:journals/ior/Brooks11}
\bibinfo{author}{J.~P. Brooks},
\newblock \bibinfo{title}{Support vector machines with the ramp loss and the
  hard margin loss},
\newblock \bibinfo{journal}{Operations Research} \bibinfo{volume}{59}
  (\bibinfo{year}{2011}) \bibinfo{pages}{467--479}.
\bibitem[{Wu and Liu(2007)}]{Wu07robusttruncated}
\bibinfo{author}{Y.~Wu}, \bibinfo{author}{Y.~Liu},
\newblock \bibinfo{title}{Robust truncated hinge loss support vector machines},
\newblock \bibinfo{journal}{Journal of the American Statistical Association}
  \bibinfo{volume}{102} (\bibinfo{year}{2007}) \bibinfo{pages}{974--983}.
\bibitem[{Zheng and Liu(2012)}]{Zheng:2012}
\bibinfo{author}{S.~Zheng}, \bibinfo{author}{W.~Liu},
\newblock \bibinfo{title}{{Functional Gradient Ascent for Probit Regression}},
\newblock \bibinfo{journal}{Pattern Recognition} \bibinfo{volume}{45}
  (\bibinfo{year}{2012}) \bibinfo{pages}{4428--4437}.
\bibitem[{Yu et~al.(2007)Yu, Zhou, Steinbac, Hand, and Steinberg}]{YZSHS2007}
\bibinfo{author}{S.~Yu}, \bibinfo{author}{Z.~H. Zhou},
  \bibinfo{author}{M.~Steinbac}, \bibinfo{author}{D.~J. Hand},
  \bibinfo{author}{D.~Steinberg},
\newblock \bibinfo{title}{Top 10 algorithms in data mining},
\newblock \bibinfo{journal}{Knowledge and Information Systems}
  \bibinfo{volume}{14} (\bibinfo{year}{2007}) \bibinfo{pages}{1--37}.
\bibitem[{Xu et~al.(2006)Xu, Crammer, and Schuurmans}]{DBLP:conf/aaai/XuCS06}
\bibinfo{author}{L.~Xu}, \bibinfo{author}{K.~Crammer},
  \bibinfo{author}{D.~Schuurmans},
\newblock \bibinfo{title}{Robust support vector machine training via convex
  outlier ablation},
\newblock in: \bibinfo{booktitle}{Proceedings of the 21st National Conference
  on Artificial Intelligence (AAAI), July 2006}, \bibinfo{publisher}{AAAI
  Press}, \bibinfo{address}{Boston, Massachusetts}, \bibinfo{year}{2006}, pp.
  \bibinfo{pages}{536--542}.
\bibitem[{McAllester and Keshet(2011)}]{conf/nips/McAllesterK11}
\bibinfo{author}{D.~A. McAllester}, \bibinfo{author}{J.~Keshet},
\newblock \bibinfo{title}{{Generalization Bounds and Consistency for Latent
  Structural Probit and Ramp Loss}},
\newblock in: \bibinfo{booktitle}{Proceddings of Advances in Neural Information
  Processing Systems (NIPS)}, \bibinfo{address}{Granada, Spain}, pp.
  \bibinfo{pages}{2205--2212}.
\bibitem[{Bache and Lichman(2013)}]{Bache+Lichman:2013}
\bibinfo{author}{K.~Bache}, \bibinfo{author}{M.~Lichman}, \bibinfo{title}{{UCI}
  machine learning repository},
  \bibinfo{howpublished}{http://archive.ics.uci.edu/ml}, \bibinfo{year}{2013}.
  \bibinfo{note}{University of California, Irvine, School of Information and
  Computer Sciences}.
\bibitem[{An and Tao(1997)}]{Pham1997}
\bibinfo{author}{L.~T.~H. An}, \bibinfo{author}{P.~D. Tao},
\newblock \bibinfo{title}{Solving a class of linearly constrained indefinite
  quadratic problems by d.c. algorithms},
\newblock \bibinfo{journal}{Journal of Global Optimization}
  \bibinfo{volume}{11} (\bibinfo{year}{1997}) \bibinfo{pages}{253--285}.
\bibitem[{Chang and Lin(2011)}]{CC01a}
\bibinfo{author}{C.-C. Chang}, \bibinfo{author}{C.-J. Lin},
  \bibinfo{title}{{LIBSVM}: A library for support vector machines},
  \bibinfo{howpublished}{http://www.csie.ntu.edu.tw/\~{}cjlin/libsvm/},
  \bibinfo{year}{2011}. \bibinfo{note}{ACM Transactions on Intelligent Systems
  and Technology}.
\bibitem[{Keerthi and Gilbert(2002)}]{Keerthi:2002}
\bibinfo{author}{S.~S. Keerthi}, \bibinfo{author}{E.~G. Gilbert},
\newblock \bibinfo{title}{Convergence of a generalized smo algorithm for svm
  classifier design},
\newblock \bibinfo{journal}{Machine Learning} \bibinfo{volume}{46}
  (\bibinfo{year}{2002}) \bibinfo{pages}{351--360}.

\end{thebibliography}
\bibliographystyle{model1-num-names}
\end{document}